\def\cvprPaperID{8001}
\def\confName{CVPR}
\def\confYear{2023}

\def\paperTitle{MAP: Multimodal Uncertainty-Aware Vision-Language Pre-training Model}

\def\authorBlock{
    Yatai Ji$^\textnormal{1}$\footnotemark[1] \qquad
    Junjie Wang$^\textnormal{2}$\footnotemark[1] \qquad
    Yuan Gong$^\textnormal{1}$ \qquad
    Lin Zhang$^\textnormal{3}$ \qquad
    Yanru Zhu$^\textnormal{1}$ \\
    Hongfa Wang$^\textnormal{4}$ \qquad
    Jiaxing Zhang$^\textnormal{3}$ \qquad
    Tetsuya Sakai$^\textnormal{2}$\footnotemark[2] \qquad
    Yujiu Yang$^\textnormal{1}$\footnotemark[2]
    \and
    $^\textnormal{1}$Tsinghua University \qquad
    $^\textnormal{2}$Waseda University \qquad
    $^\textnormal{3}$IDEA \qquad
    $^\textnormal{4}$Tencent TEG \\
    {\tt\small \{jyt21, gong-y21, zhuyr20\}@mails.tsinghua.edu.cn} \qquad {\tt\small yang.yujiu@sz.tsinghua.edu.cn} \\
    {\tt\small wjj1020181822@toki.waseda.jp} \quad {\tt\small tetsuyasakai@acm.org} \\
    {\tt\small \{zhanglin, zhangjiaxing\}@idea.edu.cn } \quad {\tt\small hongfawang@tencent.com }
}

\newif\ifreview 
\newif\ifarxiv 
\newif\ifcamera \newcommand{\cameraready}{\cameratrue}
\newif\ifrebuttal 

\cameraready

\documentclass[10pt,twocolumn,letterpaper]{article}

\ifreview \usepackage[review]{cvpr} \fi
\ifarxiv \usepackage[pagenumbers]{cvpr} \fi
\ifrebuttal \usepackage[rebuttal]{cvpr} \fi
\ifcamera \usepackage{cvpr} \fi

\usepackage{graphicx}
\usepackage{amsmath}
\usepackage{amssymb}
\usepackage{booktabs}
\usepackage{amsfonts}
\usepackage{algorithm}
\usepackage{algorithmic}
\usepackage{footnote}

\usepackage{times}
\usepackage{microtype}
\usepackage{epsfig}
\usepackage[table,xcdraw]{xcolor}
\usepackage{caption}
\usepackage{float}
\usepackage{placeins}
\usepackage{color, colortbl}
\usepackage{stfloats}
\usepackage{enumitem}
\usepackage{tabularx}
\usepackage{xstring}
\usepackage{multirow}
\usepackage{xspace}
\usepackage{url}
\usepackage{subcaption}
\usepackage{xcolor}
\usepackage[hang,flushmargin]{footmisc}

\usepackage[normalem]{ulem}
\useunder{\uline}{\ul}{}
\usepackage{booktabs}
\usepackage[switch]{lineno}
\usepackage[export]{adjustbox}
\usepackage[accsupp]{axessibility}  

\ifcamera \usepackage[accsupp]{axessibility} \fi

\ifarxiv  \fi

\newcommand{\R}[1]{{%
    \textbf{%
        \ifstrequal{#1}{1}{\textcolor{red}{R#1}}{%
        \ifstrequal{#1}{2}{\textcolor{blue}{R#1}}{%
        \ifstrequal{#1}{3}{\textcolor{magenta}{R#1}}{%
        \ifstrequal{#1}{4}{\textcolor{teal}{R#1}}{%
                           \textcolor{cyan}{R#1}%
        }}}}%
    }%
}}

\usepackage{xr-hyper}

\makeatletter
\newcommand*{\addFileDependency}[1]{
  \typeout{(#1)}
  \@addtofilelist{#1}
  \IfFileExists{#1}{}{\typeout{No file #1.}}
}

\makeatother

\usepackage[pagebackref,breaklinks,colorlinks]{hyperref}
\usepackage[capitalize]{cleveref}
\crefname{section}{Sec.}{Secs.}
\crefname{table}{Table}{Tables}
\crefname{figure}{Fig.}{Figs.}

\title{\paperTitle}
\author{\authorBlock}

\begin{document}
\maketitle

{
  \renewcommand{\thefootnote}%
    {\fnsymbol{footnote}}
  \footnotetext[1]{Equal contribution.}
  \footnotetext[2]{Corresponding Author.}
}

\begin{abstract}
Multimodal semantic understanding often has to deal with uncertainty, which means the obtained messages tend to refer to multiple targets.
Such uncertainty is problematic for our interpretation, including inter- and intra-modal uncertainty. 
Little effort has studied the modeling of this uncertainty, particularly in pre-training on unlabeled datasets and fine-tuning in task-specific downstream datasets. 
In this paper, we project the representations of all modalities as probabilistic distributions via a Probability Distribution Encoder (PDE) by utilizing sequence-level interactions. 
Compared to the existing deterministic methods, such uncertainty modeling can convey richer multimodal semantic information and more complex relationships. 
Furthermore, we integrate uncertainty modeling with popular pre-training frameworks and propose suitable pre-training tasks: Distribution-based Vision-Language Contrastive learning (D-VLC), Distribution-based Masked Language Modeling (D-MLM), and Distribution-based Image-Text Matching (D-ITM). 
The fine-tuned models are applied to challenging downstream tasks, including image-text retrieval, visual question answering, visual reasoning, and visual entailment, and achieve state-of-the-art results. 
\end{abstract}
\vspace{-0.5cm}


\section{Introduction}
\label{sec:introduction}

Precise understanding is a fundamental ability of human intelligence, whether it involves localizing objects from similar semantics or finding corresponding across multiple modalities. 
Our artificial models suppose to do the same, pinpointing exact concepts from rich multimodal semantic scenarios. 
However, this kind of precise understanding is challenging. 
Information from different modalities can present rich semantics from each other, but the resulting ambiguity and noise are also greater than the case with a single modality. 

\begin{figure}[tp]
  \includegraphics[width=0.49\textwidth]{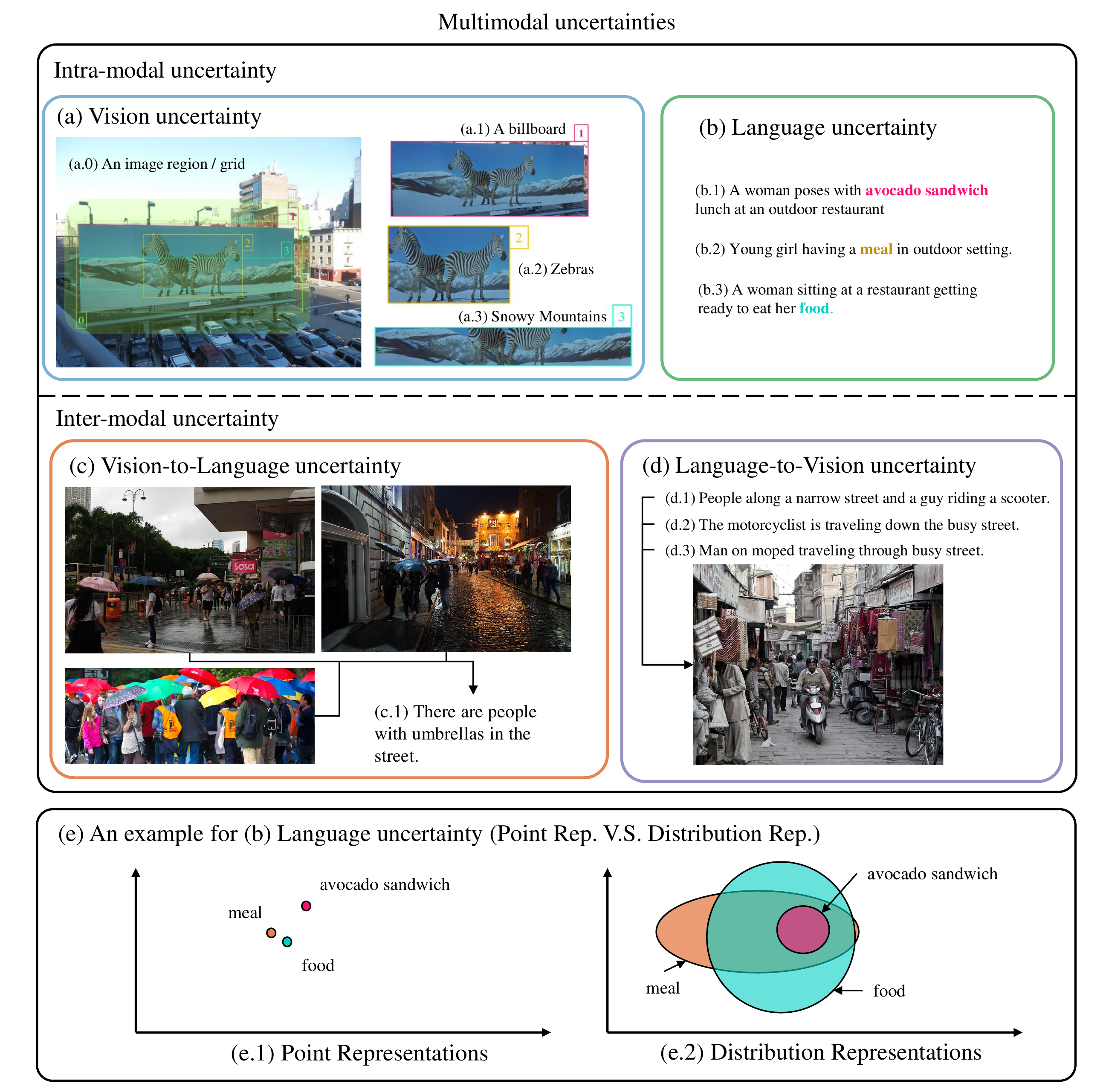}
  \caption{Multimodal uncertainties and an example for language uncertainty (b) by modeling as point representations and distribution representations. The images and text are from MSCOCO~\cite{DBLP:conf/eccv/LinMBHPRDZ14/mscoco}.}
  \label{fig:multimodal_unc}
\vspace{-0.5cm}
\end{figure}

Multimodal representation learning methods hold the promise of promoting the desired precise understanding across different modalities~\cite{Guo2019DeepMR}. 
While these methods have shown promising results, current methods face the challenge of uncertainty~\cite{DBLP:conf/cvpr/YangZ0WY21/pum,DBLP:conf/cvpr/ChunORKL21/pcme}, including within a modality and between modalities. 
Considering image (a.0) in~\cref{fig:multimodal_unc} as an example, one vision region includes multiple objects, such as a billboard, several zebras and others. 
Therefore, it is unclear which objects when mentioning this region. 
In the language domain, the complex relationships of words lead to uncertainty, such as synonymy and hyponymy. 
In~\cref{fig:multimodal_unc} (c)\&(d), the same object often has different descriptions from different modalities, such as text and images, which manifests inter-modal uncertainty. 
Instead, previous methods often neglect the uncertainty~\cite{Jing2020OvercomingLP/olp,garderes-etal-2020-conceptbert/ConceptBert,wang2021mirtt}, resulting in limited understanding ability on complicated concept hierarchies and poor prediction diversity.
Therefore, it is desirable to model such uncertainty.

Moreover, with multimodal datasets becoming more commonplace, there is a flourishing trend to implement pre-training models, particularly Vision-Language Pre-training (VLP), to support downstream applications~\cite{DBLP:conf/icml/RadfordKHRGASAM21/clip, chen2020uniter, DBLP:conf/icml/JiaYXCPPLSLD21/align, DBLP:conf/acl/XuYLBHXH20/e2e-vlp,DBLP:conf/icml/KimSK21/vilt}. 
Existing deterministic representations, however, often fail to understand uncertainty in pre-training data, as they can only express positions in semantic space and measure the relationship between targets in certainty, such as Euclidean distance. 
\textit{How can we efficiently model uncertainty in multi-modalities when dealing with pre-training models?}

Applying Gaussian distribution is one of the prominent approaches used for modeling uncertainty in the representation space~\cite{DBLP:journals/corr/VilnisM14/word2gauss, DBLP:conf/cvpr/YangZ0WY21/pum, DBLP:conf/iccv/YuLYHX19, DBLP:conf/mm/SuLSHW21}. 
In these methods, however, the obtained uncertainty depends on individual features rather than considering the whole features together, which ignores the inner connection between features. 
To exploit this connection, we implicitly model them when formulating the uncertainty with a module called Probability Distribution Encoder (PDE). 
Inspired by the self-attention mechanism~\cite{DBLP:conf/nips/VaswaniSPUJGKP17/transformer}, we further add the interaction between text tokens and image patches when constructing our distribution representations to capture more information. 
In Figure~\ref{fig:multimodal_unc} (e), we provide an example for two different types of representations to describe the language uncertainty, where the distribution representations can express richer semantic relationships than the conventional point representations. 
The distribution variance measures the uncertainty of the corresponding text. 
As a byproduct, distribution representations enable diverse generations, providing multiple reasonable predictions with random sampling. 

In this paper, we integrate this uncertainty modeling in the pre-training framework, resulting in three new tasks: Distribution-based Vision-Language Contrastive learning (D-VLC), Distribution-based Masked Language Modeling (D-MLM), and Distribution-based Image-Text Matching (D-ITM) pre-training tasks. 
All these tasks are to deal with cross-modality alignment. 
More specifically, D-VLC is to handle the coarse-grained cross-modal alignment, which measures the whole distributions to align representations from different domains. 
D-MLM and D-ITM are implemented after the fine-grained interaction between different modalities, providing the token level and overall level alignment for images and text.

Our contributions are summarized as follows:\\
1) We focus on the semantic uncertainty of multimodal understanding and propose a new module, called Probability Distribution Encoder, to frame the uncertainty in multimodal representations as Gaussian distributions.\\
2) We develop three uncertainty-aware pre-training tasks to deal with large-scale unlabeled datasets, including D-VLC, D-MLM, and D-ITM tasks. To the best of our knowledge, these are the first attempt to harness the probability distribution of representations in VLP.\\
3) We wrap the proposed pre-training tasks into an end-2-end \textbf{M}ultimodal uncertainty-\textbf{A}ware vision-language \textbf{P}re-training model, called MAP, for downstream tasks. Experiments show MAP gains State-of-The-Art (SoTA) performance.
Our code is available at https://github.com/IIGROUP/MAP.

\begin{figure*}[tp]
  \centering
  \includegraphics[width=0.96\textwidth]{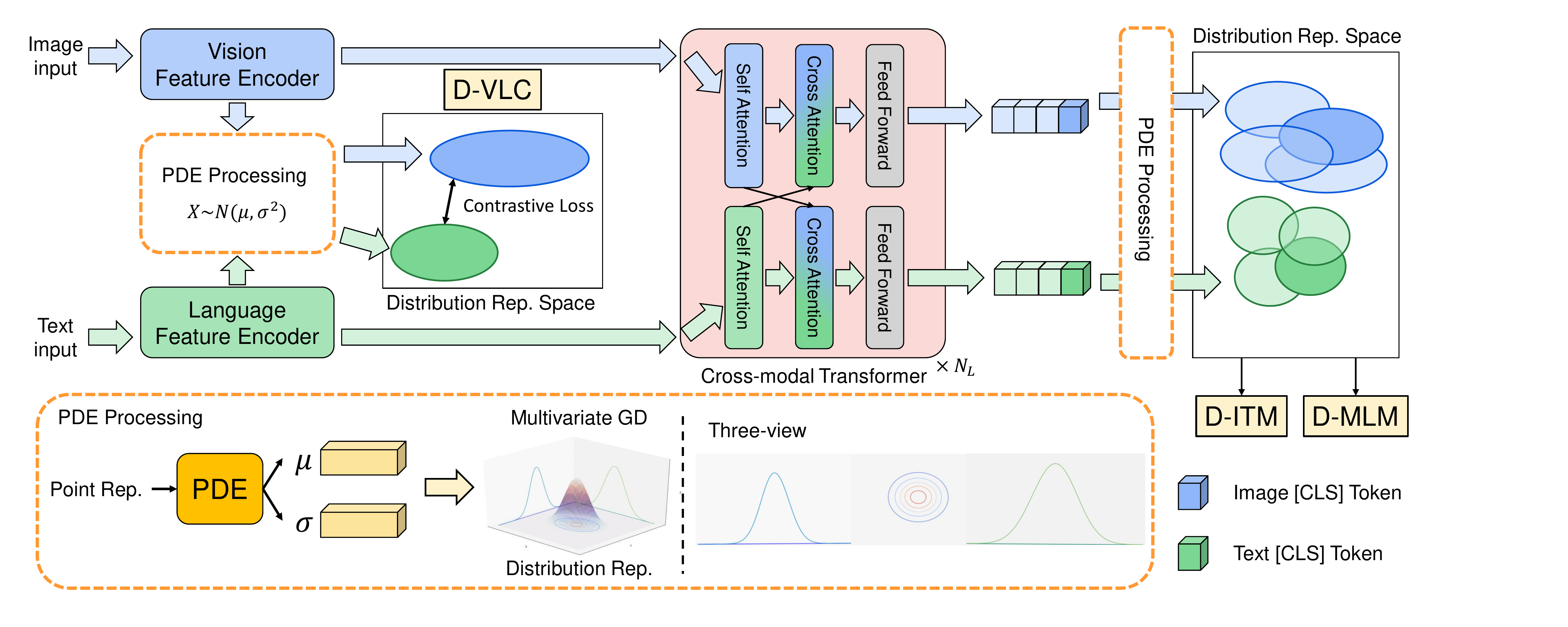}
  \caption{Pre-training model architecture and objectives of MAP. We propose PDE to model distribution representations as multivariate Gaussian Distributions (GD). ``$N_L$'' indicates the layer number of the cross-modal transformer. We perform two-dimensional Gaussian distribution as an example.}
  \label{fig:model}
\vspace{-0.5cm}
\end{figure*}

\section{Related Works}
\label{sec:related_work}

\subsection{Probability Distribution Representations}

Current popular representation learning methods extract features as point representations and focus on searching for the closest position to ground truth in high-level representation space. 
However, there is usually more than one suitable point representation, which shows the uncertainty in multiple tasks. 
To address this problem, the following researchers introduced probability distribution representations to infer diversely and improve robustness, avoiding model overfitting to one single solution. 
In the Natural Language Processing (NLP) field, multivariate Gaussian distribution was utilized to represent words~\cite{DBLP:journals/corr/VilnisM14/word2gauss} due to the powerful capability for representing the asymmetric relations among words. 
Since then, different distribution families were exploited for word representations~\cite{DBLP:conf/acl/AthiwaratkunW17, DBLP:conf/iclr/LiVZBM19}. 
In Computer Vision (CV), for modeling vision uncertainty, some researchers introduce Gaussian representations into specific tasks, such as face recognition~\cite{DBLP:conf/cvpr/ChangLCW20}, person re-identification~\cite{DBLP:conf/iccv/YuLYHX19}, 3D skeleton action representation~\cite{DBLP:conf/mm/SuLSHW21} and pose estimation~\cite{DBLP:conf/eccv/SunZCSA020}. 
For solving the long-tail problem in relation prediction, Gaussian distribution was utilized to build objects relationship in scene graph generation~\cite{DBLP:conf/cvpr/YangZ0WY21}.
Recently, constructing distributions achieved some progress to yield diverse predictions for cross-modal retrieval in multimodal field~\cite{DBLP:conf/cvpr/ChunORKL21/pcme}. 
However, those existing methods only consider the feature level to build the distributions for a whole image or sentence. 
In this work, we model not only the whole image or sentence to the distribution representations but also each token of them, such as patches and words. Furthermore, our approach learns the multimodal uncertainty from sequence-level and feature-level interactions.

\subsection{Vision-Language Pre-training (VLP)}

Inspired by the Transformer structure~\cite{DBLP:conf/nips/VaswaniSPUJGKP17/transformer} and pre-training tasks from BERT~\cite{DBLP:conf/naacl/DevlinCLT19/bert}, the recent emergence of vision-language pre-training tasks and models have been explored to learn multimodal representations. 
The main process is first to pre-train the models by exploiting auxiliary tasks to understand hidden supervision information from large-scale unlabeled data. 
Then, the pre-trained models embed real-world objects into multimodal representations. 
With effective universal multimodal representations, they can achieve good performance by fine-tuning on relatively small labeled datasets of VL downstream tasks. 
The key challenge of VLP is to design suitable pre-training objectives.
Recently, mainstream strategies include Masked Language Modeling (MLM)~\cite{DBLP:conf/cvpr/Hong0QOG21/vln-bert, DBLP:conf/iccv/KamathSLSMC21/mdeter, DBLP:conf/icml/KimSK21/vilt, huang2021seeing/soho, li2021albef}, Image-Text Matching (ITM)~\cite{DBLP:conf/cvpr/Hong0QOG21/vln-bert, DBLP:conf/icml/KimSK21/vilt, huang2021seeing/soho, li2021albef} and Vision-Language Contrastive learning (VLC)~\cite{DBLP:conf/iccv/KamathSLSMC21/mdeter, DBLP:conf/icml/JiaYXCPPLSLD21/align, li2021albef, DBLP:conf/icml/RadfordKHRGASAM21/clip}. 
MLM in VLP requires the model to predict the masked language context tokens by the rest of the language context tokens and vision context tokens. 
To understand alignment information of language context and vision context, ITM requires the model to judge whether the input of different modalities matches or not. VLC learns the similarity from inter-modal information and aligns point representations of different modalities.
However, those methods only are designed in the point representation space without considering multimodal uncertainty. 
Therefore, we propose the D-VLC, D-MLM and D-ITM to pre-train our model in the distribution representation space. 
The details will be explained in~\cref{sec:dis_based_pt}.
\vspace{-0.2cm}

\section{Approaches}
\label{sec:approaches}

In this section, we introduce our proposed PDE and the architecture of MAP (\cref{sec:model_overview}), and the overall structure is described in~\cref{fig:model}. 
The details of our proposed distribution-based pre-training tasks are presented in~\cref{sec:dis_based_pt}.
In addition, we further discuss the ethical considerations in~\cref{append:ethical}.

\subsection{Model Overview}
\label{sec:model_overview}

\subsubsection{Probability Distribution Encoder (PDE).}
\label{sec:desgin_pde}

The input features of PDE are from the point representation space of different modalities. 
To model the multimodal uncertainty, we further frame the input features as multivariate Gaussian distributions. 
Specifically, PDE predicts a mean vector ($\mu$) and a variance vector ($\sigma^2$) for each input feature. 
The mean vector represents the center position of distributions in probabilistic space, and the variance vector expresses the scope of distributions in each dimension.

\begin{figure}
  \includegraphics[width=0.49 \textwidth]{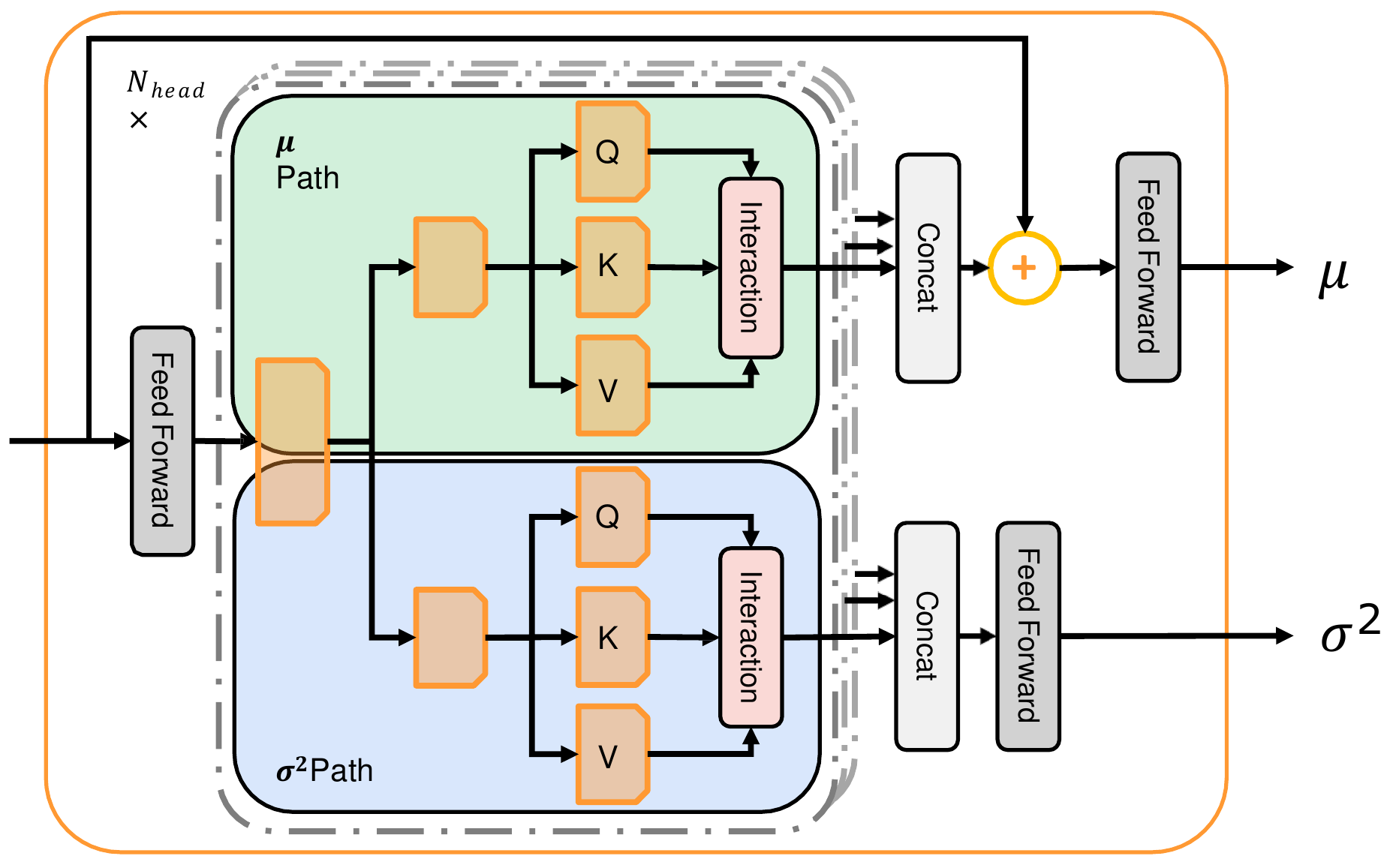}
  \caption{The architecture of Probability Distribution Encoder (PDE) block.}
  \label{fig:pde}
\vspace{-0.5cm}
\end{figure}

As shown in~\cref{fig:pde}, 
we propose a probability distribution encoder (PDE) while considering that modeling the mean and variance vectors takes feature-level and sequence-level interactions.
Specifically, Feed Forward layer is used for feature-level interactions and Multi-Head (MH) operation is responsible for sequence-level interactions. 
By applying the MH operation, the input hidden states $H\in \mathbb{R}^{T \times D}$ are split into $k$ heads, where $T$ is sequence length and $D$ is hidden size. 
In each head, we split the features and send them to two paths ($\mu$, $\sigma^2$). 
In each path, the input hidden states $H^{(i)}\in \mathbb{R}^{T \times D/{2k}}$ are projected to $Q^{(i)}$, $K^{(i)}$, $V^{(i)}$ in $i$-th head. 
As an example, the operation in the $\mu$ path is:
\begin{equation}
\small
\begin{aligned}
  &[Q_{\mu}^{(i)}, K_{\mu}^{(i)}, V_{\mu}^{(i)}] = H_{\mu}^{(i)}{W}_{qkv}\, , \\
  &Head_{\mu}^{(i)} = \operatorname{Act}\left(Q_{\mu}^{(i)} {K_{\mu}^{(i)}}^\top / \sqrt{d_k}\right)V_{\mu}^{(i)} \label{eq:activation}\, , \\
  &\operatorname{MH}_{\mu} = \operatorname{concat}_{i \in {[k]}}\big[Head_{\mu}^{(i)}\big] \; W_O\, ,
\end{aligned}
\end{equation}
where $d_k$ is set to $D/{(2k)}$. 
The weight $W_{qkv} \in \mathbb{R}^{d_k \times 3 d_k}$ is to project the inputs in the subspace of each head. 
The weight $W_O \in \mathbb{R}^{k d_k \times D}$ projects the concatenation of $k$ head results to the output space. 
The ``$\operatorname{Act}$'' includes an activation function and a normalization function for considering sequence-level interaction. 
The $\sigma^2$ path is similar to the $\mu$ path. 
Since the input point representation correlates with the mean vector, an add operation is employed to learn the mean vector. 
The motivations of design choices are in~\cref{sec:ablation_activation_function}. 
After PDE, each vision or language token is represented as a gaussian distribution in high-dimension probabilistic space.

\subsubsection{Feature Extraction.}
To extract features, we utilize an image encoder and a language encoder. 
In detail, we employ CLIP-ViT~\cite{DBLP:conf/icml/RadfordKHRGASAM21/clip} as the image encoder and RoBERTa-Base~\cite{DBLP:journals/corr/abs-1907-11692/roberta} as the language encoder. 
An image is encoded as an patch feature sequence $\{v_{\tt [CLS]},v_1, \dots ,v_N\}$ , where $v_{\tt [CLS]}$ is the overall feature. 
Moreover, the input text is embedded into a sequence of tokens $\{w_{\tt [CLS]},w_1, \dots,w_M\}$.

\subsubsection{Cross-modal Transformer.}

Recently, there are two main types of the multimodel transformer to fuse the different modalities:  single-stream~\cite{DBLP:conf/iclr/SuZCLLWD20/vlbert,chen2020uniter, DBLP:conf/cvpr/ZhangLHY0WCG21/vinvl} and dual-stream~\cite{DBLP:conf/emnlp/TanB19/lxmert,DBLP:conf/nips/LuBPL19/vilbert, li2021albef} models. 

In our method, the image patch sequences are much longer than text sequences, making the weights of vision features too large to compute the attention scores together. 
To address this issue, we choose the dual-stream module with two transformer branches, where self-attention scores are calculated separately. 

As shown in~\cref{fig:model}, the main structure has $N_L$ layers of cross-modal encoders. 
Each encoder mainly consists of two Self-Attention (SA) blocks and two Cross-Attention (CA) blocks. 
In the SA block of each modality, query, key and value vectors are all linearly projected from vision or language features. 
In the vision-to-language cross-attention block of $i$-th layer, query vectors represent language feature $T_{i}'$ after the self-attention block, and key/value vectors denote vision feature $I_i'$. 
By employing the Multi-Head Attention (MHA) operation, the CA block enables language features to learn visual information across modalities. 
The language-to-vision CA block is similar to the vision-to-language one. 
The workflow of $i$-th layer encoder with SA and CA is as follows:
\begin{equation}
\small
\begin{aligned}
SA_{\text{vision}}: I_i'&=\operatorname{MHA}(I_{i-1},I_{i-1},I_{i-1}), \\
SA_{\text{language}}: T_i'&=\operatorname{MHA}(T_{i-1},T_{i-1},T_{i-1}), \\
CA_{\text{vision}}: I_i&=\operatorname{MHA}(I_i',T_{i}',T_{i}'), \\
CA_{\text{language}}: T_i&=\operatorname{MHA}(T_i',I_{i}',I_{i}'). \\
\end{aligned}
\label{cross-att encoder}
\end{equation}

For the overall structure design of MAP, we apply PDEs after feature extractors and cross-modal transformer, respectively. 
PDE after the feature extractor learns unimodal distribution representations to conduct the D-VLC pre-training task. 
PDE at the end of MAP is responsible for D-MLM, D-ITM and downstream tasks.

\subsection{Distribution-based Pre-Training Tasks}
\label{sec:dis_based_pt}
In order to learn the multimodal uncertainty in common sense, we pre-train our model with distribution-based pre-training tasks on large-scale datasets.

\subsubsection{Coarse-grained Pre-training.}

We propose Distribution-based Vision-Language Contrastive Learning, called~\textbf{D-VLC}, to realize coarse-grained semantic alignment of overall unimodal distributional representations before fusion.
We compute the 2-Wasserstein distance~\cite{mallasto2017learning/2wassertein, kantorovich1960mathematical/2wassertein, kantorovich2006translocation/2wassertein} to measure the distance between multivariate Gaussian distributions. 
For two Gaussian distributions $\mathcal{N}(\mu_1,\Sigma_1)$ and $\mathcal{N}(\mu_2,\Sigma_2)$, their 2-Wasserstein distance is defined as:
\begin{equation}
\small
D_{2W}=||\mu_1-\mu_2||_2^2+\operatorname{Tr}(\Sigma_1+\Sigma_2-2(\Sigma_1^{1/2}\Sigma_2\Sigma_1^{1/2})^{1/2}).
\end{equation}

In our modeled distributions, $\Sigma_1$ and $\Sigma_2$ are both diagonal matrices, which indicates $\Sigma_1^{1/2}\Sigma_2\Sigma_1^{1/2}=\Sigma_1\Sigma_2$. 
The above formula can be rewritten as:

\begin{equation}
\small
\begin{aligned}
D_{2W}&=||\mu_1-\mu_2||_2^2+\operatorname{Tr}((\Sigma_1^{1/2}-\Sigma_2^{1/2})^2) \\
&=||\mu_1-\mu_2||_2^2+||\sigma_1-\sigma_2||_2^2\, ,
\end{aligned}
\end{equation}
where $\sigma$ refers to a standard deviation vector. 
The overall unimodal features denote the distribution representations of ${\tt [CLS]}$ from the PDEs following single-modal feature extractors. 
The similarity between an image and text is given by:
\begin{equation}
\small
    s(I,T)=a\cdot D_{2W}(v_{\tt [CLS]}, w_{\tt [CLS]})+b\, ,
\label{eq:similarity}
\end{equation}
where $a$ is a negative scale factor since similarity is inversely proportional to the distance, and $b$ is a shift value. 
For $N$ image-text pairs in a batch, there are $N$ positive matched samples and $N(N-1)$ negative samples. 
We use InfoNCE loss as follows: 
\begin{equation}
\small
\begin{aligned}
\mathcal{L}_{NCE}^{I2T}(i)=-\operatorname{log}\frac{\operatorname{exp}(s(I_i,T_i)/\tau)}{\sum_{n=1}^N\operatorname{exp}(s(I_i,T_n)/\tau)}\, , \\
\mathcal{L}_{NCE}^{T2I}(i)=-\operatorname{log}\frac{\operatorname{exp}(s(T_i,I_i)/\tau)}{\sum_{n=1}^N\operatorname{exp}(s(T_i,I_n)/\tau)}\, ,
\end{aligned}
\end{equation}
where $\tau$ is a learned temperature parameter. 
The above are summed as D-VLC loss $\mathcal{L}_{D\mbox{-}VLC}$. 

\subsubsection{Fine-grained Pre-training.}

After the cross-modal transformer with fine-grained interaction on each token of different modalities, our proposed Distribution-based Masked Language Modeling (D-MLM) and Distribution-based Image Text Matching (D-ITM) can assist the model in learning fine-grained cross-modal alignment.

\textbf{D-MLM} requires the model to predict the masked words by understanding the text with an image. 
Specifically, the input text tokens are replaced by ${\tt [MASK]}$ with a probability of 15\% (Details in~\cref{append:d_mlm}). 
For conducting classification in the word list to predict the original words, we sample the point vectors from distribution representations. 
D-MLM minimizes a Cross-Entropy (CE) loss for $\mu$ point and other sample point vectors:
\begin{equation}
\small
    \mathcal{L}_{D\mbox{-}MLM}=\frac{1}{K+1}(\operatorname{CE}(\phi(\mu),y)+\sum_{i=1}^K\operatorname{CE}(\phi(z^{(i)}), y)),
\end{equation}
where $K$ is the sample number from gaussian distributions and $y$ serves as a masked word label. 
$\mu$ is a mean vector and $z^{(i)}$ refers to stochastic sample point vectors; then, they are fed to MLM classifier $\phi$. 
During the inference process, the final output is the mean pooling of all samples' prediction results:
\begin{equation}
\small
    P=\frac{1}{K+1}(\phi(\mu)+\sum_{i=1}^K\phi(z^{(i)})).    
\end{equation}

\textbf{D-ITM} provides a binary classification that predicts whether a pair of image-text is matched or not. 
Specifically, we sample the point vectors from $w_{\tt CLS}$ distributions of vision and language features and concatenate them as the fusion features to generate the prediction.
\begin{equation}
\small
\begin{aligned}
    \mathcal{L}_{D\mbox{-}ITM} &= \frac{1}{K+1}(\operatorname{CE}(\phi(\operatorname{concat}[v_{\mu}, w_{\mu}]),y) \\
    &+\sum_{i=1}^K\operatorname{CE}(\phi(\operatorname{concat}[v^{(i)}, w^{(i)}]), y)),
\end{aligned}
\end{equation}
where $v_\mu$, $t_\mu$ are mean vectors of vision and language ${\tt [CLS]}$ distributions. 
$v^{(i)}$, $w^{(i)}$ are sample points and $\phi$ is the D-ITM classifier. 
The image-text pairs in the datasets serve as positive examples, and negative examples are constructed by randomly replacing images or text descriptions.

However, random sampling increases training difficulty. 
When the model is trained only with the aforementioned losses, it will lead to variance collapse. 
Since all sample point vectors will converge to the optimal position, the distribution representations eventually degenerate into point representations, resulting in losing the ability to learn multimodal uncertainty. 
Therefore, we append a regularization loss to prevent the uncertainty level of distributions is lower than a certain threshold:
\begin{equation}
\small
    \mathcal{L}_{reg}=\operatorname{max}(0,\gamma -h(\mathcal{N}(\mu,\sigma^2))),
\label{eq:loss_reg}
\end{equation}
where $\gamma$ is a set threshold, which affects the uncertainty level of distributions. 
$h(\mathcal{N}(\mu,\sigma^2))$ is the entropy of a multivariate Gaussian distribution, which should be defined as:
\begin{equation}
\small
    h(\mathcal{N}(\mu, \Sigma))=\frac{1}{2}\operatorname{log}(\operatorname{det}(2\pi e\Sigma)),
\label{eq:entropy_ori}
\end{equation}
where $\Sigma$ is the covariance matrix, which is a diagonal matrix in our method. 
Therefore, the diagonal vector of $\Sigma$ is $\sigma^2$ and the~\cref{eq:entropy_ori} can be transformed to: 
\begin{equation}
\small
\begin{aligned}
h(\mathcal{N}(\mu,\sigma^2))&=\frac{1}{2}\sum_{i=1}^d\operatorname{log}(2\pi e \cdot \sigma_i^2) \\
&=\frac{d}{2}(\operatorname{log}(2\pi)+1)+\sum_{i=1}^d\operatorname{log}\sigma_i\, ,
\end{aligned}
\end{equation}
where $d$ is the feature dimension.

Note that the sampling operation for $\mathcal{N}(\mu,\sigma^2)$ causes the problem of preventing gradients from propagating back. 
By applying the \emph{reparameterization trick}~\cite{DBLP:journals/corr/KingmaW13/VAE}, we first sample a random noise $\epsilon$ from standard normal distributions, instead of directly sampling from $\mathcal{N}(\mu,\sigma^2)$:
\begin{equation}
\small
    z=\mu+\sigma \epsilon, \,\,\,\,\, \epsilon\sim~\mathcal{N}(0,I).
    \label{eq:reparameterization}
\end{equation}
After~\cref{eq:reparameterization}, the output $z$ obeys the predicted distributions from the PDE. 
Therefore, we can separate the calculations of the mean and standard deviation from the sampling operation and they are trainable.

\subsubsection{Training Objectives.}

During pre-training phase, the model will propagate forward three times at one step with conducting D-MLM, D-ITM and D-VLC tasks separately. 
Therefore, the full pre-training objective is given by:
\begin{equation}
\small
    \mathcal{L}_{pre}=\mathcal{L}_{D\mbox{-}MLM}+\mathcal{L}_{D\mbox{-}ITM}+\mathcal{L}_{D\mbox{-}VLC}+\alpha \mathcal{L}_{reg}\, ,
\label{eq:full_pt_loss}
\end{equation}
where $\alpha$ is its weight.

\section{Experiments}
\label{sec:experiments}

\subsection{Experimental settings}
\label{sec:implementation_details}

By following a popular setting~\cite{Dou_2022_CVPR/meter}, we set all hidden feature sizes as 768, and the head number as 12 in MHA. 
Unless otherwise specified, the layer number ($N_L$) of the cross-modal transformer is set to $6$. 
As for data processing, we resize and crop each image into the size of $384\times 384$. 
The size $P$ of the image patch is $16$. 
And the maximum length of input text dealt is set to $50$.
In PDE, the head number $k$ is set to 6 and the default ``Act'' function in~\cref{eq:activation} is Softmax. 

For pre-training, we pre-train our model with D-MLM, D-ITM and D-VLC. 
The pre-training datasets include MSCOCO~\cite{DBLP:conf/eccv/LinMBHPRDZ14/mscoco}, Visual Genome (VG)~\cite{DBLP:journals/ijcv/KrishnaZGJHKCKL17/vg}, SBU~\cite{DBLP:conf/nips/OrdonezKB11/sbu} and Conceptual Captions (CC-3M)~\cite{DBLP:conf/acl/SoricutDSG18/cc3m}. 
Specifically, we resize and crop each image into the size of $288\times 288$. 
Please find more pre-training and fine-tuning details in~\cref{append:exp_settings}.

In all experiments, we employ the randomized Tukey HSD p-values and effect sizes based on one-way ANOVA~\cite{sakai2018laboratory/hsd} to support the statistical significance of all results (Please refer to~\cref{append:hsd} for more details).

\subsection{Results of VL Downstream Tasks}
\label{sec:main_results}

In this section, we apply our pre-trained MAP on the following $4$ VL downstream tasks with $5$ widely-used datasets. 
Image retrieval task (MSCOCO~\cite{DBLP:conf/eccv/LinMBHPRDZ14/mscoco} and Flickr30K~\cite{DBLP:conf/iccv/PlummerWCCHL15/f30k}) aims to understand the multimodal uncertainty which results from the multiplicity of concepts in images and text. 
This is similar to the objective of our uncertainty modeling in nature. 
Meanwhile, visual question answering (VQA2.0~\cite{balanced_vqa_v2/vqa_v2}), visual reasoning (NLVR2~\cite{DBLP:conf/acl/SuhrZZZBA19/nlvr2}) and visual entailment (SNLI-VE~\cite{DBLP:journals/corr/abs-1811-10582/snli-ve}) implicitly perform ambiguous semantics in unimodal and cross-modal items. 
Therefore, we further evaluate our MAP on the aforementioned tasks to varying the effectiveness and generalization ability of uncertainty modeling. 
For fair experimental environments, we group previous models with different sizes of pre-training datasets. 
Please find more details of the datasets, model descriptions and additional experiments in~\cref{append:exp_details}.

\begin{table*}[t]
\small
\centering
\begin{adjustbox}{max width=\textwidth}
\begin{tabular}{l|cccccc|cccccc}
\toprule
                        & \multicolumn{6}{c|}{MSCOCO (5K test set)}                                                            & \multicolumn{6}{c}{Flickr30K (1K test set)}                                                         \\
\multirow{-2}{*}{Model} & IR@1           & IR@5           & IR@10          & TR@1           & TR@5           & TR@10          & IR@1           & IR@5           & IR@10          & TR@1           & TR@5           & TR@10          \\ \midrule
\rowcolor[HTML]{EFEFEF} 
\multicolumn{13}{l}{\textit{\cellcolor[HTML]{EFEFEF}Group 2: Pre-training datasets include $> 10$M images}}                                                                                                                                  \\ \midrule
\rowcolor[HTML]{EFEFEF} 
ALBEF (14M)~\cite{li2021albef}             & 60.7           & 84.3           & 90.5           & 77.6           & 94.3           & 97.2           & 85.6           & 97.5           & 98.9           & 95.9           & 99.8           & 100.0          \\ \midrule
\multicolumn{13}{l}{\textit{Group 1: Pre-training datasets include $< 10$M images}}                                                                                                                                                             \\ \midrule
UNITER-Large~\cite{chen2020uniter}            & 52.9          & 79.9          & 88.0          & 65.7         & 88.6          & 93.8          & 75.6          & 94.1          & 96.8          & 87.3          & 98.0          & 99.2          \\
VILLA-Large~\cite{DBLP:conf/nips/Gan0LZ0020/villa}             & -              & -              & -              & -              & -              & -              & 76.3          & 94.2          & 96.8          & 87.9          & 97.5          & 98.8          \\
UNIMO-Large~\cite{DBLP:conf/acl/LiGNXLL0020/unimo}             & -              & -              & -              & -              & -              & -              & 78.0          & 94.2          & 97.1          & 89.4          & 98.9          & {\ul 99.8}    \\
VinVL-Large~\cite{DBLP:conf/cvpr/ZhangLHY0WCG21/vinvl}             & 58.8     & {\ul 83.5}     & {\ul 90.3}     & 75.4           & 92.9           & 96.2           & -              & -              & -              & -              & -              & -              \\ \midrule
ViLT~\cite{DBLP:conf/icml/KimSK21/vilt}                    & 42.7           & 72.9           & 83.1           & 61.5           & 86.3           & 92.7           & 64.4           & 88.7           & 93.8           & 83.5           & 96.7           & 98.6           \\
UNITER-Base~\cite{chen2020uniter}             & 50.3          & 78.5          & 87.2          & 64.4          & 87.4          & 93.1          & 72.5          & 92.4          & 96.8          & 85.9          & 97.1          & 98.8          \\
ALBEF (4M)~\cite{li2021albef}              & 56.8          & 81.5          & 89.2          & 73.1          & 91.4          & 96.0          & 82.8          & 96.7          & 98.4          & 94.3          & 99.4          & {\ul 99.8}    \\
TCL~\cite{DBLP:conf/cvpr/YangDTXCCZCH22/tcl} & {\ul 59.0} & 83.2 & 89.9 & 75.6 & 92.8 & 96.7 & \textbf{84.0} & {\ul 96.7} & {\ul 98.5} & \textbf{94.9} & {\ul 99.5} & {\ul 99.8} \\
METER~\cite{Dou_2022_CVPR/meter}          & 57.1          & 82.7          & 90.1          & {\ul 76.2}    & {\ul 93.2}    & {\ul 96.8}    & 82.2    & 96.3    & 98.4    & {\ul 94.3}    & \textbf{99.6} & \textbf{99.9} \\
MAP  (ours)              & \textbf{60.9} & \textbf{86.2} & \textbf{93.1} & \textbf{79.3} & \textbf{94.8} & \textbf{97.6} & {\ul 83.8} & \textbf{97.2} & \textbf{98.7} & \textbf{94.9} & {\ul 99.5}    & {\ul 99.8} \\
\bottomrule
\end{tabular}
\end{adjustbox}
\caption{An overall comparison with SoTA models on fine-tuned image-text retrieval tasks. The best scores are in \textbf{bold} and the second best scores are \underline{underlined}.}
\label{table:result_ir}
\vspace{-0.3cm}
\end{table*}

\begin{table}[t]
\small
\centering
\begin{adjustbox}{max width=0.5\textwidth}
\begin{tabular}{l|c|cc|cc}
\toprule
\multirow{2}{*}{Model} & VQA2.0         & \multicolumn{2}{c|}{NLVR2}       & \multicolumn{2}{c}{SNLI-VE}     \\
                       & test-dev       & dev            & test-p         & val            & test           \\ \midrule
\rowcolor[HTML]{EFEFEF} 
\multicolumn{6}{l}{\textit{\cellcolor[HTML]{EFEFEF}Group 2: Pre-training datasets include \textgreater 10M images (Base size)}}            \\ \midrule
\rowcolor[HTML]{EFEFEF} 
ALBEF (14M)~\cite{li2021albef}             & 75.84          & 81.72          & 81.77          & 84.20          & 84.15          \\
\rowcolor[HTML]{EFEFEF} 
SimVLM-Base~\cite{DBLP:journals/corr/abs-2108-10904/simvlm}             & 77.87          & 82.55          & 83.14          & 80.80          & 80.91          \\ \midrule
\multicolumn{6}{l}{\textit{Group 1: Pre-training datasets include \textless 10M images (Base size)}}                                      \\ \midrule
ViLT~\cite{DBLP:conf/icml/KimSK21/vilt}                   & 71.26          & 75.70          & 76.13          & -              & -              \\
UNITER-Base~\cite{chen2020uniter}           & 72.70          & 77.18          & 77.85          & 78.59          & 78.28          \\
OSCAR-Base~\cite{DBLP:conf/eccv/Li0LZHZWH0WCG20/oscar}             & 73.16          & 78.07          & 78.36          & -              & -              \\
UNIMO-Base~\cite{DBLP:conf/acl/LiGNXLL0020/unimo}             & 73.79          & -              & -              & 80.00          & 79.10          \\
ALBEF (4M)~\cite{li2021albef}             & 74.54          & 80.24          & 80.50          & 80.14          & 80.30          \\
VinVL-Base~\cite{DBLP:conf/cvpr/ZhangLHY0WCG21/vinvl}             & 75.95          & 82.05          & 83.08          & -              & -              \\

VLMo-Base~\cite{DBLP:journals/corr/abs-2111-02358/vlmo}              & 76.64          & {\ul 82.77}    & {\ul 83.34}    & -              & -              \\
METER~\cite{Dou_2022_CVPR/meter}         & {\ul 77.68}    & 82.33          & 83.05          & {\ul 80.86}    & {\ul 81.19}    \\
MAP (ours)             & \textbf{78.03} & \textbf{83.30} & \textbf{83.48} & \textbf{81.40} & \textbf{81.39} \\
\bottomrule
\end{tabular}
\end{adjustbox}
\caption{An overall comparison with SoTA models on visual question answering, visual reasoning, visual entailment tasks. The best scores are in \textbf{bold} and the second best scores are \underline{underlined}.}
\label{table:result_vqa}
\vspace{-0.3cm}
\end{table}

\subsubsection{Evaluation on Image-Text Retrieval}

As shown in~\cref{table:result_ir}, our MAP achieves the best performance on MSCOCO and gains either the best or the second-best scores on Flickr30K. 
Specially, while ALBEF has specially-designed objectives for retrieval, the MAP also outperforms ALBEF ($14$M pre-training images) in all metrics on the MSCOCO retrieval task. 
The results show the effectiveness and advantages of uncertainty modeling. 
For the Flickr30K dataset, our MAP achieves the best performance or only about $0.1$ point behind the best score. 
PCME also utilizes probabilistic distribution representations to conduct retrieval task, and we show the comparison in~\cref{append:pcme}.

\subsubsection{Evaluation on VQA2.0, NVLR2, and SNLI-VE}

As shown in~\cref{table:result_vqa}, our MAP outperforms the previous SoTA models in Group 1. 
Compared to VLMo-Base, the MAP improves $0.53$ points on NLVR2 dev. 
Our model brings the $+0.35$ points improvement on VQA2.0 test-dev and $+0.54$ points performance gains on SNLI-VE val over METER. 
Notably, MAP outperforms SimVLM-Base ($1.8$B pre-training images) in all tasks, which further demonstrates the effectiveness of uncertainty modeling.

\begin{table}[t]
\small
\centering
\begin{tabular}{lccccc}
\toprule
\multicolumn{1}{l|}{}            & \multicolumn{1}{c|}{VQA2.0}         & \multicolumn{2}{c|}{SNLI-VE}                         & \multicolumn{2}{c}{NLVR2}       \\
\multicolumn{1}{l|}{}            & \multicolumn{1}{c|}{test-dev}       & val            & \multicolumn{1}{c|}{test}         & dev            & test-p           \\ \midrule
\multicolumn{6}{l}{\textit{Random initialization}}                                                                                                                      \\ \midrule
\multicolumn{1}{l|}{MAP w/o PDE} & \multicolumn{1}{c|}{72.09}          & 75.91          & \multicolumn{1}{c|}{76.28}          & 50.86          & 51.07          \\ 
\multicolumn{1}{l|}{MAP}         & \multicolumn{1}{c|}{73.35}          & 76.67          & \multicolumn{1}{c|}{76.86}          & 51.12          & 51.07          \\ \midrule
\multicolumn{6}{l}{\textit{Pretained on MSCOCO}}                                                                                                                  \\ \midrule
\multicolumn{1}{l|}{MAP w/o PDE} & \multicolumn{1}{c|}{74.57}          & 79.42          & \multicolumn{1}{c|}{79.84}          & 77.72          & 79.31          \\
\multicolumn{1}{l|}{MAP}         & \multicolumn{1}{c|}{\textbf{75.01}} & \textbf{80.05} & \multicolumn{1}{c|}{\textbf{80.31}} & \textbf{78.96} & \textbf{79.64} \\ \bottomrule
\end{tabular}
\caption{The effectiveness of probability distribution representations on VL downstream tasks. For ``MAP w/o PDE'', we train a new model without PDE to conduct the experiments. Pre-trained methods for MAP: D-MLM, D-ITM. Pre-trained methods for MAP w/o PDE: MLM, ITM.}
\label{table:pde_or_not}
\vspace{-0.3cm}
\end{table}

\subsection{Ablation Studies}
\label{sec:ablation_studies}

\subsubsection{How do the probability distribution representations affect VL downstream task?}
As shown in~\cref{table:pde_or_not}, applying PDE helps the model to achieve a better performance, which matters significantly in VL downstream tasks. 
In both random and pre-trained weights initialization cases, distribution representations gain a better capability of VL understanding than the point representations, which is because the distribution representations can express richer semantics by learning multimodal uncertainty.

\subsubsection{ How does the structure of PDE behave?} 
\label{sec:ablation_activation_function}

We remove the sequence-level interaction in PDE and call it ``MLP only'' (MultiLayer Perceptron), which is the common method in the previous works~\cite{DBLP:conf/cvpr/ChunORKL21/pcme, DBLP:conf/cvpr/YangZ0WY21/pum, DBLP:conf/iccv/YuLYHX19}. 
Table~\ref{table:pde_struc} shows that PDE (Softmax)  outperforms ``MLP only'' in VQA2.0, which benefits from the sequence-level information. 
Moreover, we design several candidate activation functions: ReLU, ReLU$^2$, Sigmoid, and Softmax. 
Notably, ``MLP only'' outperforms ``ReLU'' and ``ReLU$^2$'', which demonstrates that it is important to consider how to design the sequence-level interaction. 
The function Sigmoid projects the input values between 0 to 1, which smoothly assigns weights between different tokens. 
The function Softmax outperforms the others in VQA2.0, which implies that Softmax is suitable to express the correlation between tokens. 
Therefore, we set Softmax as the default activation function in sequence-level interaction.

\begin{table}[]
\small
\centering
\begin{tabular}{cl|c}
\toprule
\multicolumn{2}{l|}{Structure}                              & VQA2.0 (test-dev) \\ \midrule
\multicolumn{2}{l|}{MLP only}                               & 72.01            \\ \midrule
\multicolumn{1}{c|}{\multirow{4}{*}{PDE}} & ReLU+Normal     & 69.70            \\
\multicolumn{1}{c|}{}                     & ReLU$^2$+Normal & 70.53            \\
\multicolumn{1}{c|}{}                     & Sigmoid+Normal  & 73.34            \\
\multicolumn{1}{c|}{}                     & Softmax         & \textbf{73.35}            \\ \bottomrule
\end{tabular}
\caption{Effect of different structures of PDE. We explore the different designs of ``Act'' in Equation~\ref{eq:activation}. Normal denotes the normalization operation.}
\label{table:pde_struc}
\end{table}

\begin{table}[]
\small
\centering
\begin{tabular}{l|c|c|c}
\toprule
\multirow{2}{*}{Training strategies}        & VQA2.0   & SNLI-VE & NLVR2 \\
            & test-dev & test-p  & test  \\ \midrule
Random Initialization   & 73.35    & 76.86   & 51.07 \\
D-MLM, D-ITM     & 75.01    & 80.31   & \textbf{79.64} \\
D-MLM, D-VLC     & 75.06    & 80.12   & 77.90  \\
D-ITM, D-VLC     & 71.02     & 78.54    & 73.64  \\
D-MLM, D-ITM, D-VLC & \textbf{75.16}    & \textbf{80.39}   & 79.47 \\ \bottomrule
\end{tabular}
\caption{The effect of distribution-based pre-training tasks. We pre-train the model on the MSCOCO dataset.}
\label{table:pre-training}
\end{table}

\subsubsection{What is the performance of different pre-training objectives?}
\cref{table:pre-training} presents that different choices of pre-training tasks affect the VL downstream tasks performance. 
According to the chart, results without D-MLM pre-training are the worst in all pre-training strategies, which means D-MLM plays the most important role in pre-training. 
Both D-VLC and D-ITM assist the model in learning semantic similarity between vision and language modality.
In VQA2.0, D-VLC makes a larger improvement than D-ITM, whereas D-ITM is more effective than D-VLC in SNLI-VE and NLVR2.

\begin{table}[t]
\small
\centering
\begin{tabular}{c|c|c}
\toprule
Layer Number & Random Initializing & Pre-training \\ \midrule
2            & 72.71               & 73.78        \\
4            & 73.32               & 74.73        \\
6            & \textbf{73.35}               & 75.16        \\
8            & 73.31               & \textbf{75.26}        \\ \bottomrule
\end{tabular}
\caption{The effect of different layer numbers in the cross-modal transformer on VQA2.0.}
\label{table:cross_trans_layers}
\vspace{-0.4cm}
\end{table}

\subsubsection{Does the number of layers of cross-modal transformer matter?}

As shown in~\cref{table:cross_trans_layers}, we explore the effect of layer number on VQA2.0 by random initializing or pre-training with D-MLM, D-ITM, D-VLC on the MSCOCO dataset. 
By random initializing, the model with six layers achieves the best performance, which encounters a bottleneck. 
After pre-training, the model with eight layers makes little progress from six layers, which implies that pre-training helps the model break the aforementioned bottleneck of parameters. 
The reason perhaps is that pre-training with large-scale data alleviates the problem of over-fitting by more parameters. 
Moreover, as the layer number decreases, the effect of pre-training will reduce due to the model's limited learning capacity.

\begin{figure}[tp]
  \centering
  \includegraphics[width=0.49\textwidth]{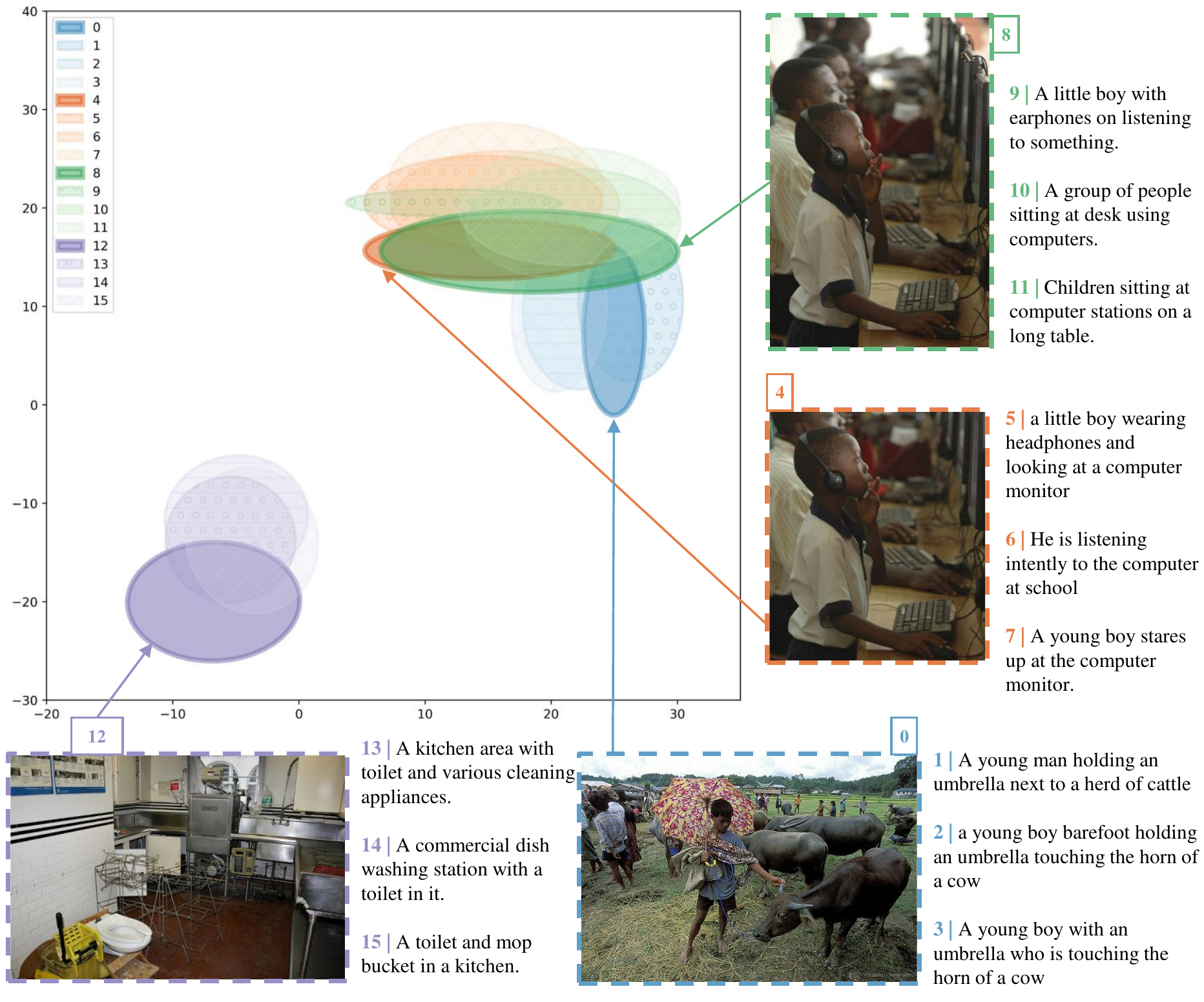}
  \caption{Visualization of the distribution representations from pre-trained MAP\protect\footnotemark[2]. Each 2D Gaussian distribution is represented as an ellipse with 95\% confidence. The labels of images and related captions are in the legend.}
  \label{fig:visualization}
 \vspace{-0.4cm}
\end{figure}

\footnotetext[1]{The images and related captions come from MSCOCO dataset~\cite{DBLP:conf/eccv/LinMBHPRDZ14/mscoco}.}

\subsection{Uncertainty Modeling Analysis}
\label{sec:analysis}

\noindent\textbf{Visualization.} To perform visualization analysis for distribution representations from pre-trained MAP, we conduct 2D toy experiments for the images and related descriptions. 
\cref{fig:visualization} shows the behaviors of the distribution representations, which present that distributions with similar semantics are clustered together. 
The shapes of images and related descriptions are similar, and the ellipses are closed, showing that the images and text cover similar meanings. 
For example, since image ``4'' is a part of image ``8'', ellipse ``8'' almost includes all regions of ellipse ``4''. 
The intersection of ellipses (images ``0'', ``4'', ``8'' and their corresponding captions) might indicate ``a young boy'' in images and text. 
Similar behaviors of our MAP can be found in more visualizations in~\cref{append:add_examples_visualization}.
Intuitively, as shown in visualization results, uncertainty modeling facilitates the model to express rich semantic information and complex relationships. 

\begin{figure}[tp]
  \centering
  \includegraphics[width=0.45\textwidth]{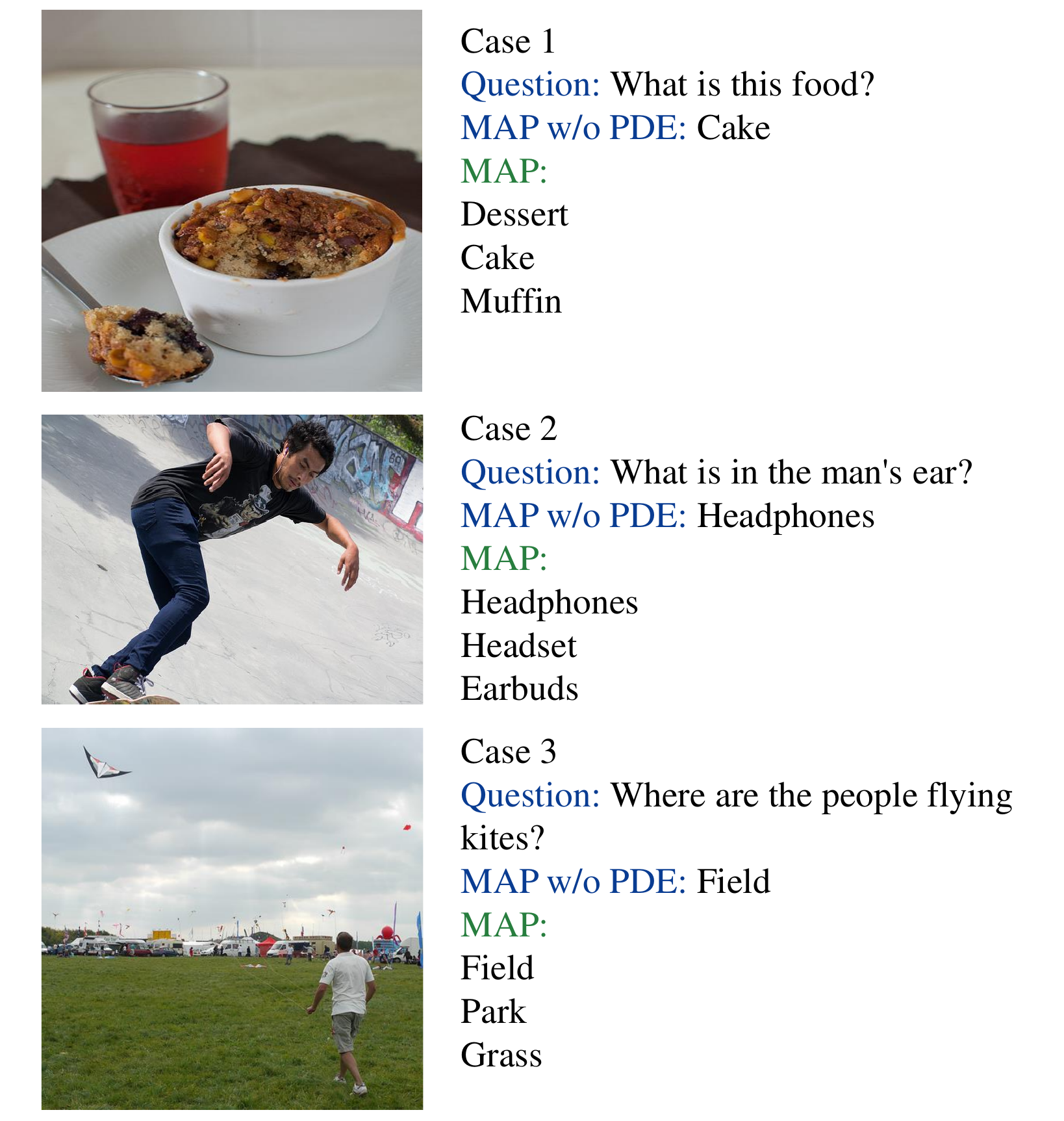}
  \caption{Predictions sampled from the distribution representations\protect\footnotemark[3].}
  \label{fig:diversity_examples}
\vspace{-0.4cm}
\end{figure}

\noindent\textbf{Cases for diverse predictions.} 
Semantic uncertainty is ubiquitous in multimodal tasks. 
For multimodal understanding tasks such as VQA, an advantage of uncertainty modeling is that multiple predictions can be sampled from distribution representations, which provides diversity. 
Consider case $3$ in~\cref{fig:diversity_examples}, MAP can learn multiple plausible answers (field, park and grass) from the distribution representations, which is close to our real world. 
In contrast, the point representations from MAP without PDE always generate one answer ignoring other possible expressions. 
Moreover, distribution representations can also help other multimodal tasks, such as image captioning, to generate several suitable captions, which benefit from diverse correspondences caused by uncertainty modeling.

\section{Conclusions}
\label{sec:conclusion}

In this work, we focus on the multimodal uncertainty in real-world objects by modeling this onto probability distributions. 
By considering sequence-level and feature-level interactions, we proposed a Probability Distribution Encoder (PDE) to gain distribution representations for different modalities. 
Our experiments showed that distribution representations are beneficial for the VL downstream tasks. 
In addition, uncertainty modeling facilitates diverse predictions. 
To learn multimodal uncertainty in large-scale data, we designed three new pre-training tasks (D-MLM, D-ITM and D-VLC). 
Furthermore, we propose an end-to-end \textbf{M}ultimodal uncertainty-\textbf{A}ware vision-language \textbf{P}re-training model (MAP) to obtain generic distribution representations. 
We demonstrate the effectiveness of the proposed MAP on several VL downstream tasks empirically. 
In the future, we will explore more distribution subspaces and experiments on larger datasets.

\section*{Acknowledgements}
This work was partly supported by the National Natural Science Foundation of China (Grant No. 61991450) and the Shenzhen Science and Technology Program (JSGG20220831093004008; ZDSYS20200811142605016).

\newpage
{\small
\bibliographystyle{ieee_fullname}
\bibliography{egbib}
}

\clearpage
\appendix
\label{sec:appendix}

\section*{Appendix}

\section{Architecture Details}
\label{append:archi_details}

\subsection{Multi-head Attention}
\label{append:mha}
By following the instruction from Transformer~\cite{DBLP:conf/nips/VaswaniSPUJGKP17/transformer}, the $Q$, $K$, $V$ are computed from input hidden states $H \in \mathbb{R}^{T \times D}$ and $H' \in \mathbb{R}^{T' \times D}$. 
The two input matrices consist of respectively $T$ and $T'$ tokens of $d$ dimensions each. 
The transformation is as follows:
\begin{equation}
\small
\begin{aligned}
Q &= HW_Q 
& W_Q &\in \mathbb{R}^{D \times d_k}\,,\\
K &= H'W_K
& W_K &\in \mathbb{R}^{D \times d_k}\,,\\
V &= H'W_V
& W_V &\in \mathbb{R}^{D \times d_k}\,.
\end{aligned}
\end{equation}

An attention map is computed by the pairwise similarity between two tokens from $H$ and $H'$.

\begin{equation}
\small
\operatorname{Attention}(Q,K,V) = \operatorname{softmax}\left(QK^\top / \sqrt{d_k}\right)V,
\end{equation}
After splitting $k$ heads from $H$ and $H'$, the Multi-Head Attention (MHA) is concatenated from the outputs by running $k$ attention operations. 
The same calculations of $Q$, $K$, $V$ are conducted in each $i \in {[k]}$ head to form $Q^{(i)}$, $K^{(i)}$, $V^{(i)}$.

\begin{equation}
\small
\begin{aligned}
  Head^{(i)} &= \operatorname{Attention}(Q^{(i)}, K^{(i)}, V^{(i)})\,,\\
  \operatorname{MHA}(Q, K, V) &=\operatorname{concat}_{i \in {[k]}}\big[Head^{(i)}\big] \; W_O\,,
\end{aligned}
\end{equation}
where the weight $W_O \in \mathbb{R}^{k d_k \times D}$ projects the concatenation of $k$ head results to the output space $D$ with the same dimension of the inputs. 
In our models, we set $d_k = D  / k$. 
The other contents in the transformer block, such as MLP Block and residual connection, follow the instructions of Transformer~\cite{DBLP:conf/nips/VaswaniSPUJGKP17/transformer}. 
$D$ is set to $768$ and $k$ is set to $12$ in our experiments. The detailed experiment settings are presented in the open-sourced codes.

\subsection{Multi-head operation in PDE}
\label{append:interaction_pde}
The operation is similar to the aforementioned multi-head attention in~\ref{append:mha}. In this operation, the input hidden states $H\in \mathbb{R}^{T \times D}$ are split into $k$ heads, where $T$ is sequence length and $D$ is hidden size. 
In each head, we split the features and send them to two paths ($\mu$, $\sigma^2$). 
The operation in the $\sigma^2$ path is followed:

\begin{equation}
\small
\begin{aligned}
  &[Q_{\sigma^2}^{(i)}, K_{\sigma^2}^{(i)}, V_{\sigma^2}^{(i)}] = H{W}_{qkv}\, , \\
  &Head_{\sigma^2}^{(i)} = \operatorname{Act}\left(Q_{\sigma^2}^{(i)} {K_{\sigma^2}^{(i)}}^\top / \sqrt{d_k}\right)V_{\sigma^2}^{(i)}\,, \\
  &\operatorname{MH}_{\sigma^2}(Q_{\sigma^2}^{(i)}\,, K_{\sigma^2}^{(i)}, V_{\sigma^2}^{(i)}) = \operatorname{concat}_{i \in {[k]}}\big[Head_{\sigma^2}^{(i)}\big] \; W_O \,,
\end{aligned}
\end{equation}
where $d_k$ is set to $D/{(2k)}$. 
The weight $W_{qkv} \in \mathbb{R}^{d_k \times 3 d_k}$ projects the inputs to the sub-space in each head. 
The weight $W_O \in \mathbb{R}^{k d_k \times D}$ projects the concatenation of $k$ head results to the output space. 
The ``$\operatorname{Act}$'' is an activation function and normalization function for considering sequence-level interaction. 
Moreover, the ``$\operatorname{MH}$'' is the multi-head operation. 
On the $\sigma^2$ path, since the predicted vector has negative values from the activation function, PDE is expected to predict $\operatorname{log}\sigma$. 
After a simple $\operatorname{exp}$ operation, variance vectors are obtained. 
There are some candidate activation functions are considered: ReLU, ReLU$^2$, Sigmoid, and Softmax. 
Unless otherwise specified, the function Softmax is employed in PDE.

\subsection{D-MLM settings}
\label{append:d_mlm}
Masked Language Modeling (MLM) is first utilized as a pre-training strategy of BERT~\cite{DBLP:conf/naacl/DevlinCLT19/bert} to predict masked words, which enhances the ability of contextual modeling. 
In multimodal pre-training, the missing words are reconstructed with retained text and information from another modality. 
The model can correctly identify the entity relationships between text and images, learning cross-model semantic alignment. 
Following the settings from several multimodal models~\cite{Dou_2022_CVPR/meter, DBLP:conf/icml/KimSK21/vilt}, the model randomly covers the text tokens with a probability of $15\%$, where $80\%$ tokens are replaced with ${\tt [MASK]}$ token, $10\%$ tokens are replaced with other random words, and $10\%$ tokens remain unchanged.

\begin{table}[t]
\small
\centering
\begin{tabular}{l|cc}
\toprule
Dataset   & \#Images & \#Text \\ \midrule
Flickr30K~\cite{DBLP:conf/iccv/PlummerWCCHL15/f30k}   & $29$K  & $145$K \\
GQA~\cite{DBLP:conf/cvpr/HudsonM19/gqa}   & $79$K  & $1$M \\ \midrule
MSCOCO~\cite{DBLP:conf/eccv/LinMBHPRDZ14/mscoco} & $113$K   & $567$K \\
VG~\cite{DBLP:journals/ijcv/KrishnaZGJHKCKL17/vg}   & $108$K   & $5.4$M \\
SBU~\cite{DBLP:conf/nips/OrdonezKB11/sbu}  & $875$K   & $875$K \\
CC-3M~\cite{DBLP:conf/acl/SoricutDSG18/cc3m} & $3.1$M     & $3.1$M   \\
CC-12M~\cite{DBLP:conf/cvpr/ChangpinyoSDS21/cc12m} & $12$M & $12$M  \\
ALIGN~\cite{DBLP:conf/icml/JiaYXCPPLSLD21/align} & $1.8$B  & $1.8$B \\ 
\bottomrule
\end{tabular}
\caption{Details of pre-training datasets in Table~\ref{table:all_models}.}
\label{table:pt_dataaset_details}
\end{table}

\begin{figure*}[]
  \includegraphics[width=0.98\textwidth]{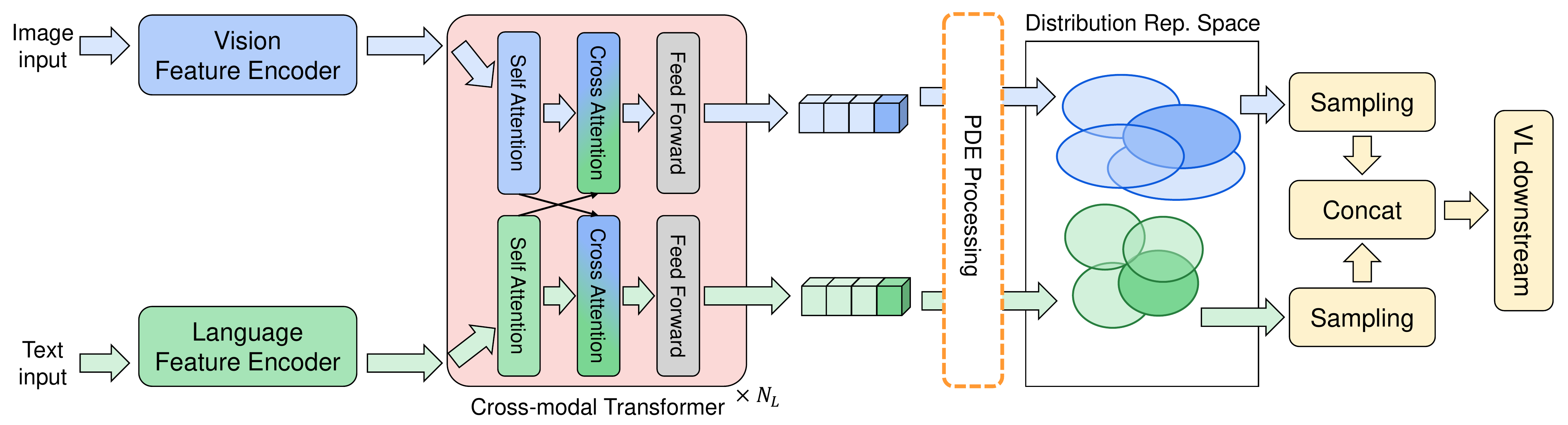}
  \caption{Fine-tuning MAP on different VL downstream tasks.}
  \label{fig:ft_cap}
\end{figure*}

\begin{table*}[t]
\small
\centering
\begin{tabular}{l|c|c|c}
\toprule
Model            & Paper   & Pre-training Datasets          & Model size  \\ \midrule
\multicolumn{4}{l}{\textit{Pre-training datasets include $> 10$M images}} \\ \midrule
ALBEF (14M)      & \cite{li2021albef}        & MSCOCO, VG, CC-3M, SBU, CC-12M  & Base        \\
SimVLM-base      & \cite{DBLP:journals/corr/abs-2108-10904/simvlm}        & ALIGN                          & Base        \\ \midrule
\multicolumn{4}{l}{\textit{Pre-training datasets include $< 10$M images}}    \\ \midrule
UNITER-Large     & \cite{chen2020uniter}        & MSCOCO, VG, CC-3M, SBU           & Large       \\
VILLA-Large      & \cite{DBLP:conf/nips/Gan0LZ0020/villa}        & MSCOCO, VG, CC-3M, SBU           & Large       \\
UNIMO-Large      & \cite{DBLP:conf/acl/LiGNXLL0020/unimo}        & MSCOCO, VG, CC-3M, SBU           & Large       \\
VinVL-large      & \cite{DBLP:conf/cvpr/ZhangLHY0WCG21/vinvl}        & MSCOCO, CC-3M, SBU, F30k, GQA    & Large       \\ \midrule
ViLT             & \cite{DBLP:conf/icml/KimSK21/vilt}        & MSCOCO, VG, CC-3M, SBU           & Base        \\
UNITER -Base     & \cite{chen2020uniter}        & MSCOCO, VG, CC-3M, SBU           & Base        \\
OSCAR-Base       & \cite{DBLP:conf/eccv/Li0LZHZWH0WCG20/oscar}        & MSCOCO, VG, CC-3M, SBU           & Base        \\
UNIMO-Base       & \cite{DBLP:conf/acl/LiGNXLL0020/unimo}        & MSCOCO, VG, CC-3M, SBU           & Base        \\
ALBEF (4M)       & \cite{li2021albef}        & MSCOCO, VG, CC-3M, SBU           & Base        \\
VLMo-Base        & \cite{DBLP:journals/corr/abs-2111-02358/vlmo}        & MSCOCO, VG, CC-3M, SBU           & Base        \\
TCL       & \cite{DBLP:conf/cvpr/YangDTXCCZCH22/tcl}        & MSCOCO, VG, CC-3M, SBU           & Base        \\
METER   & \cite{Dou_2022_CVPR/meter}        & MSCOCO, VG, CC-3M, SBU           & Base        \\
\bottomrule
\end{tabular}
\caption{Details of all models in Table~\ref{table:result_ir} and~\ref{table:result_vqa}. }
\label{table:all_models}
\end{table*}

\section{Experiment Details}
\label{append:exp_details}

\subsection{Experimental settings}
\label{append:exp_settings}

Our experiments are conducted on $8$ NVIDIA A100 GPUs. 
For usual settings in all experiments, we adopt the AdamW optimizer. 
The learning rate is warmed up first and then decayed linearly. 
When sampling point vectors from distribution representations, the sample number $K$ is set to $5$. 
In the pre-training phase, the model is trained for $100$K steps with a batch size of $4,096$. 
The learning rate of feature extractors is set to $1e-5$. 
Cross-modal transformer and PDE's learning rates are both $5e-5$.

For pre-training details, We pre-train our model with D-MLM, D-ITM and D-VLC. 
In Equation~\ref{eq:similarity} of D-VLC, $a$ is set to $-0.005$ and $b$ is set to $6$. 
In the full loss formula Equation~\ref{eq:full_pt_loss}, $\alpha$ is equal to $0.01$. 
For the regularization loss of distributions in Eq.~\ref{eq:loss_reg}, the threshold $\gamma=300$.

Table~\ref{table:pt_dataaset_details} reports the statistic of images and text of the pre-training datasets in Table~\ref{table:all_models}, which includes the pre-training datasets of all referenced models. 
Those datasets are constructed by combining public datasets. 
However, a substantial portion of the image URLs in datasets might be inaccessible now, which makes the number of images less than the statistic.

\subsection{Fine-tuning details}
\label{append:ft_phase}

The illustration of fine-tuning MAP on the VL downstream tasks is shown in Figure~\ref{fig:ft_cap}. 
For different downstream tasks, we just design a simple classifier for understanding tasks. 
We first sample the point vectors from distribution representations of ${\tt [CLS]}$. 
Then we concatenate point representations from different modalities as global features to conduct classification and apply average pooling operation to all samples' results. 
The model MAP is trained for $10$ epochs. 
The learning rates of feature extractors, Cross-modal transformer and PDE are $5e-6$, $2.5e-5$, and $2e-4$. 
In future work, we would like to try applying MAP to do several generation tasks by designing a simple decoder.

\subsection{Comparison details}
\label{append:comp}

We summarize all referenced modes with model size and pre-training datasets in Table~\ref{table:all_models}. 
The reported scores in Table~\ref{table:result_ir} and~\ref{table:result_vqa} come from their papers. 
As described in Section~\ref{sec:main_results}, we introduce the definition of model size~\cite{DBLP:conf/cvpr/ZhangLHY0WCG21/vinvl}. 
In detail, considering model parameter efficiency, the model size of Vision Language Pre-training (VLP) models can be categorized into at least $3$ size: Small, Base, and Large. 
(1) ``Small'' indicates the small models prior to the transformer-based VLP models. 
(2) ``Base'' indicates the VLP models with similar size to BERT-Base~\cite{DBLP:conf/naacl/DevlinCLT19/bert}. 
(3) ``Large'' is the VLP model with a similar size to BERT-Large. 
Furthermore, the details of pre-training datasets are presented in Table~\ref{table:pt_dataaset_details}.

\subsection{VL downstream tasks}
\label{append:vl_down}

\subsubsection{Visual Question Answering}

Given an image and a corresponding question, VQA2.0~\cite{balanced_vqa_v2/vqa_v2} is the task of providing a correct answer to the question.

\subsubsection{NLVR2} 
The NLVR2~\cite{DBLP:conf/acl/SuhrZZZBA19/nlvr2} task requires the system to judge whether the corresponding relationship between the description and two images is consistent.

\subsubsection{SNLI-VE} 
SNLI-VE~\cite{DBLP:journals/corr/abs-1811-10582/snli-ve} task requires understanding three categories of relationships between images and text, which are entailment, neutral or contradiction.

\subsubsection{Image-Text Retrieval} MSCOCO~\cite{DBLP:conf/eccv/LinMBHPRDZ14/mscoco} and Filkr30K~\cite{DBLP:conf/iccv/PlummerWCCHL15/f30k} includes two tasks: Image-to-Text retrieval task and Text-to-Image retrieval task. 
Both tasks require the model to rank the images or text by computing the image-text similarity scores. 
In detail, we utilize the Karpathy \& Fei-Fei $5$K MSCOCO test set and Filkr30K test set and then report the top-$K$ retrieval results. 

\begin{table*}[t]
\small
\centering
\begin{tabular}{llccc}
\toprule
Run1        & Run2      & VQA2.0             & SNLI-VE            & NLVR2              \\ \midrule
rand\_point & rand\_MAP & $p<0.001$ ($-0.193$) & $p<0.001$ ($-0.183$) & $p<0.001$ ($0.267$)  \\
rand\_point & pt\_point & $p<0.001$ ($-0.211$) & $p<0.001$ ($0.028$)  & $p<0.001$ ($-0.017$) \\
rand\_point & pt\_MAP   & $p<0.001$ ($-0.052$) & $p<0.001$ ($0.098$)  & $p<0.001$ ($0.367$)  \\
rand\_MAP   & pt\_point & $p<0.001$ ($-0.018$) & $p<0.001$ ($0.211$)  & $p<0.001$ ($-0.284$) \\
rand\_MAP   & pt\_MAP   & $p<0.001$ ($0.141$)  & $p<0.001$ ($0.280$)  & $p<0.001$ ($0.100$)  \\
pt\_point   & pt\_MAP   & $p<0.001$ ($0.159$)  & $p<0.001$ ($0.070$)  & $p<0.001$ ($0.384$)  \\ \bottomrule
\end{tabular}
\caption{Statistical significance calculated by Randomized Tukey HSD tests for Table~\ref{table:pde_or_not} after 1,000 trials. $p$-value and (effect size) for different tasks.}
\label{table:hsd_pde_or_not}
\end{table*}

\begin{table*}[t]
\small
\centering
\begin{tabular}{lcccc}
\toprule
         & MLP                & ReLU               & ReLU$^2$           & Sigmod            \\ \midrule
ReLU     & $p<0.001$ ($0.385$)  & -                  & -                  & -                 \\
ReLU$^2$ & $p<0.001$ ($0.287$)  & $p<0.001$ ($-0.098$) & -                  & -                 \\
Sigmod   & $p<0.001$ ($0.162$)  & $p<0.001$ ($-0.223$) & $p<0.001$ ($-0.125$) & -                 \\
PDE      & $p<0.001$ ($-0.151$) & $p<0.001$ ($0.234$)  & $p<0.001$ ($0.136$)  & $p<0.001$ ($0.011$) \\ \bottomrule
\end{tabular}
\caption{Statistical significance calculated by Randomized Tukey HSD tests for Table~\ref{table:pde_struc} after 1,000 trials. $p$-value and (effect size).}
\label{table:hsd_pde_struc}
\end{table*}

\begin{table*}[t]
\small
\centering
\begin{adjustbox}{max width=\textwidth}
\begin{tabular}{lccccccc}
\toprule
        & rand\_2            & rand\_4            & rand\_6            & rand\_8           & pt\_2             & pt\_4             & pt\_6             \\ \midrule
rand\_4 & $p<0.001$ ($-0.103$) & -                  & -                  & -                 & -                 & -                 & -                 \\
rand\_6 & $p<0.001$ ($-0.134$) & $p<0.001$ ($-0.031$) & -                  & -                 & -                 & -                 & -                 \\
rand\_8 & $p<0.001$ ($-0.150$) & $p<0.001$ ($-0.047$) & $p<0.001$ ($-0.016$) & -                 & -                 & -                 & -                 \\
pt\_2   & $p<0.001$ ($-0.007$) & $p<0.001$ ($0.035$)  & $p<0.001$ ($0.066$)  & $p<0.001$ ($0.082$) & -                 & -                 & -                 \\
pt\_4   & $p=0.005$ ($0.004$)  & $p<0.001$ ($0.107$)  & $p<0.001$ ($0.138$)  & $p<0.001$ ($0.154$) & $p<0.001$ ($0.072$) & -                 & -                 \\
pt\_6   & $p<0.001$ ($0.059$)  & $p<0.001$ ($0.161$)  & $p<0.001$ ($0.192$)  & $p<0.001$ ($0.208$) & $p<0.001$ ($0.126$) & $p<0.001$ ($0.054$) & -                 \\
pt\_8   & $p<0.001$ ($0.070$)  & $p<0.001$ ($0.173$)  & $p<0.001$ ($0.204$)  & $p<0.001$ ($0.220$) & $p<0.001$ ($0.138$) & $p<0.001$ ($0.066$) & $p<0.001$ ($0.012$) \\ \bottomrule
\end{tabular}
\end{adjustbox}
\caption{Statistical significance calculated by Randomized Tukey HSD tests for Table~\ref{table:cross_trans_layers} after 1,000 trials. $p$-value and (effect size).}
\label{table:hsd_cross_trans_layers}
\end{table*}

\begin{table*}[t]
\small
\centering
\begin{tabular}{llccc}
\toprule
Run1     & Run2          & VQA2.0             & SNLI-VE            & NLVR2              \\ \midrule
rand     & MLM\_ITM      & $p<0.001$ ($0.142$)  & $p<0.001$ ($0.301$)  & $p<0.001$ ($0.099$)  \\
rand     & MLM\_VLC      & $p<0.001$ ($0.149$)  & $p<0.001$ ($0.264$)  & $p<0.001$ ($0.047$)  \\
rand     & ITM\_VLC      & $p<0.001$ ($-0.624$) & $p<0.001$ ($0.588$)  & $p<0.001$ ($0.122$)  \\
rand     & MLM\_ITM\_VLC & $p<0.001$ ($0.195$)  & $p<0.001$ ($0.079$)  & $p<0.001$ ($0.202$)  \\
MLM\_ITM & MLM\_VLC      & $p<0.001$ ($0.007$)  & $p<0.001$ ($-0.038$) & $p<0.001$ ($-0.052$) \\
MLM\_ITM & ITM\_VLC      & $p<0.001$ ($-0.765$) & $p<0.001$ ($0.286$)  & $p<0.001$ ($0.023$)  \\
MLM\_ITM & MLM\_ITM\_VLC & $p<0.001$ ($0.138$)  & $p<0.001$ ($-0.222$) & $p<0.001$ ($0.103$)  \\
MLM\_VLC & ITM\_VLC      & $p<0.001$ ($0.053$)  & $p<0.001$ ($0.324$)  & $p<0.001$ ($0.075$)  \\
MLM\_VLC & MLM\_ITM\_VLC & $p<0.001$ ($-0.772$) & $p<0.001$ ($-0.185$) & $p<0.001$ ($0.155$)  \\
MLM\_VLC & MLM\_ITM\_VLC & $p<0.001$ ($0.818$)  & $p<0.001$ ($-0.509$) & $p<0.001$ ($0.080$)  \\ \bottomrule
\end{tabular}
\caption{Statistical significance calculated by Randomized Tukey HSD tests for Table~\ref{table:pre-training} after 1,000 trials. $p$-value and (effect size). MLM, ITM, VLC and rand indicate D-MLM, D-ITM, D-VLC and random initialization respectively.}
\label{table:hsd_pre-training}
\end{table*}

\begin{table}[t]
\small
\centering
\begin{tabular}{l|c}
\toprule
Model          & VQA2.0 (test-dev) \\ \midrule
ViLBERT~\cite{DBLP:conf/nips/LuBPL19/vilbert}       & 68.93            \\
MCAN~\cite{DBLP:conf/cvpr/Yu0CT019/mcan}          & 70.63            \\
UNITER~\cite{chen2020uniter}         & 67.03            \\
METER-swin~\cite{Dou_2022_CVPR/meter}     & 72.38            \\
METER-clip-vit~\cite{Dou_2022_CVPR/meter} & 71.75            \\ \midrule
MAP (ours)      & \textbf{73.35}            \\ \bottomrule
\end{tabular}
\caption{Evaluation on VQA2.0 of models with random initialization.}
\label{table:vqa_random}
\end{table}

\subsection{Additional results for random initialized MAP}

To examine the effectiveness of MAP without extra data, we compare MAP on the popular VL understanding task VQA2.0, with the existing reported methods. 
As shown in Table~\ref{table:vqa_random}, MAP achieves a SOTA performance on VQA2.0 among the existing methods without extra data. 
It shows that PDE can bring multimodal uncertainty knowledge to the models without transferring from large-scale pre-training datasets.

\subsection{Comparison between MAP and PCME}
\label{append:pcme}
PCME~\cite{DBLP:conf/cvpr/ChunORKL21/pcme} is a dual-tower architecture for retrieval, which uses soft contrastive loss with sampled points from distributions. In contrast, Our contrastive loss is based on 2W distance, which directly measures multiple distributions. From a quantization perspective, thanks to pre-training, MAP has a significant boost over PCME. On COCO5k test set, PCME's scores are 44.2/31.9 on I2T/T2I, MAP's scores are 79.3/60.9. 

\subsection{P-value based on Randomized Tukey HSD tests}
\label{append:hsd}

In all experiments for our implemented models, $p$-values were obtained using the randomized Tukey HSD test~\cite{sakai2018laboratory/hsd}. 
The names of runs refer to the related tables. 
In the experiments, we evaluate the test split in all tasks. 
Table~\ref{table:hsd_pde_struc} reports the Randomized Tukey HSD tests for Table~\ref{table:pde_struc}. 
In details of Table~\ref{table:hsd_pde_or_not}, the name of runs follows the rule: W\_M, where $\text{W} \in \{ \text{rand}, \text{pt} \}$ is random initialization or pre-training and $\text{M} \in \{ \text{point}, \text{MAP} \}$ is utilizing ``MAP w/o PDE'' or ``MAP''. 
Similarly, Table~\ref{table:hsd_cross_trans_layers} is the conducted tests for Table~\ref{table:cross_trans_layers} with name rules: W\_L, where $\text{W} \in \{ \text{rand}, \text{pt} \}$ and $\text{L} \in \{ 2, 4, 6, 8 \}$ is the number of layers of cross-modal transformer in MAP. The tests for Table~\ref{table:pre-training} are shown in Table~\ref{table:hsd_pre-training}.

\begin{figure*}[t]
  \includegraphics[width=0.90\textwidth]{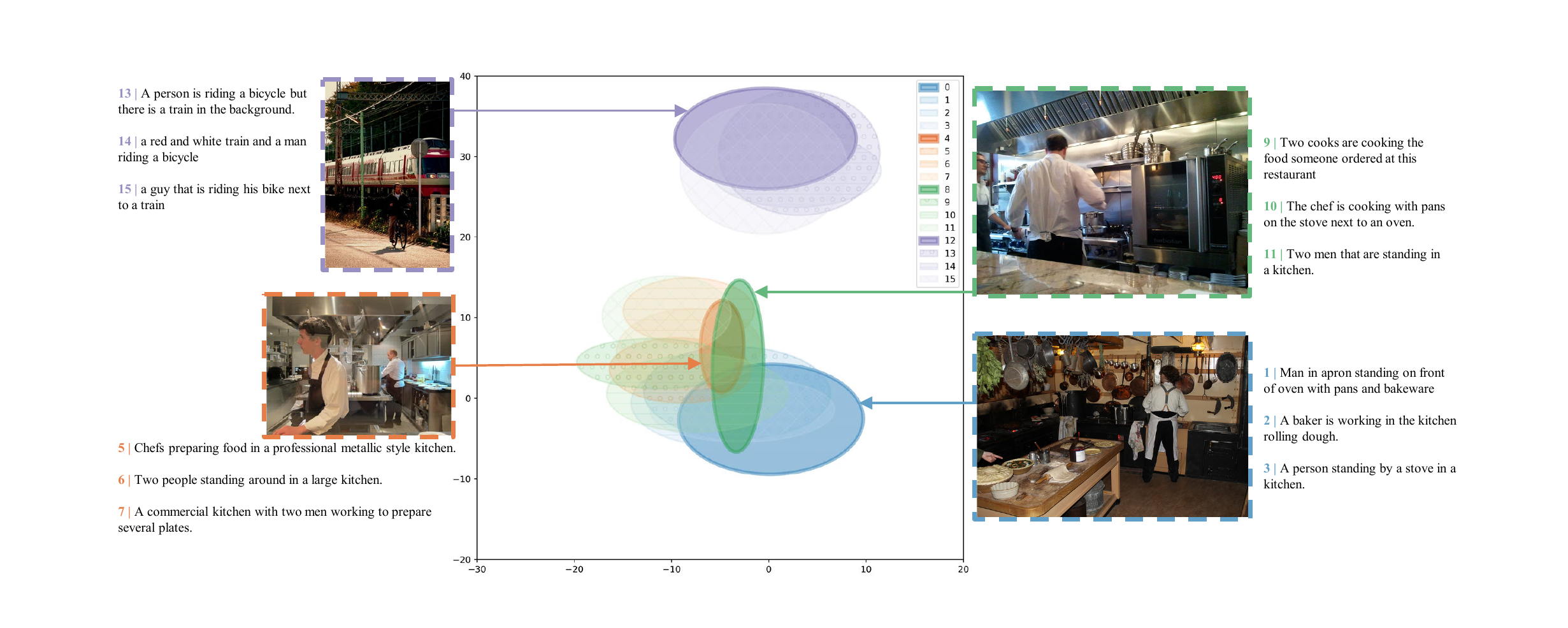}
  \caption{Additional example 1. There are some images and captions of ``chef'', ``kitchen'', ``person'', ``bike'' and so on.}
  \label{fig:visualization_others_1}
\end{figure*}

\begin{figure*}[t]
  \includegraphics[width=0.90\textwidth]{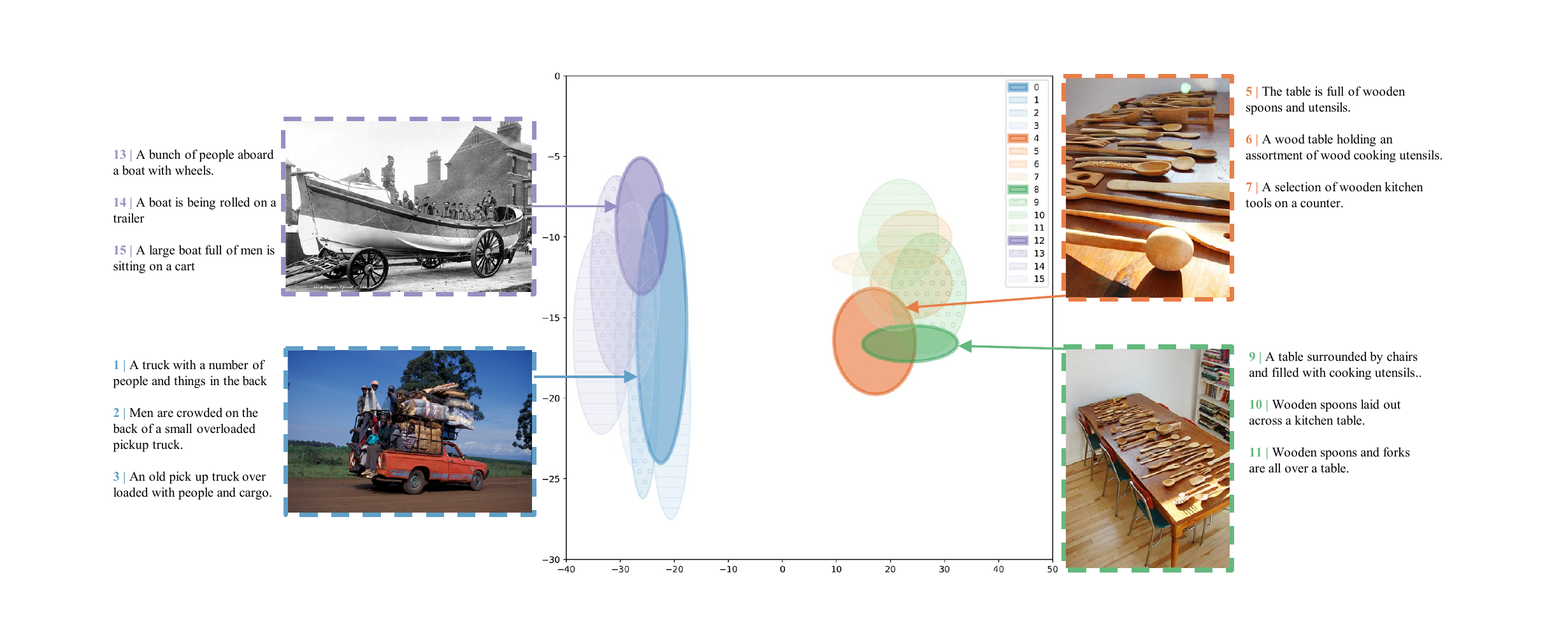}
  \caption{Additional example 2. There are some images and captions of ``utensils'', ``people'', ``truck'', ``table'' and so on.}
  \label{fig:visualization_others_2}
\end{figure*}

\begin{figure*}[t]
  \includegraphics[width=0.90\textwidth]{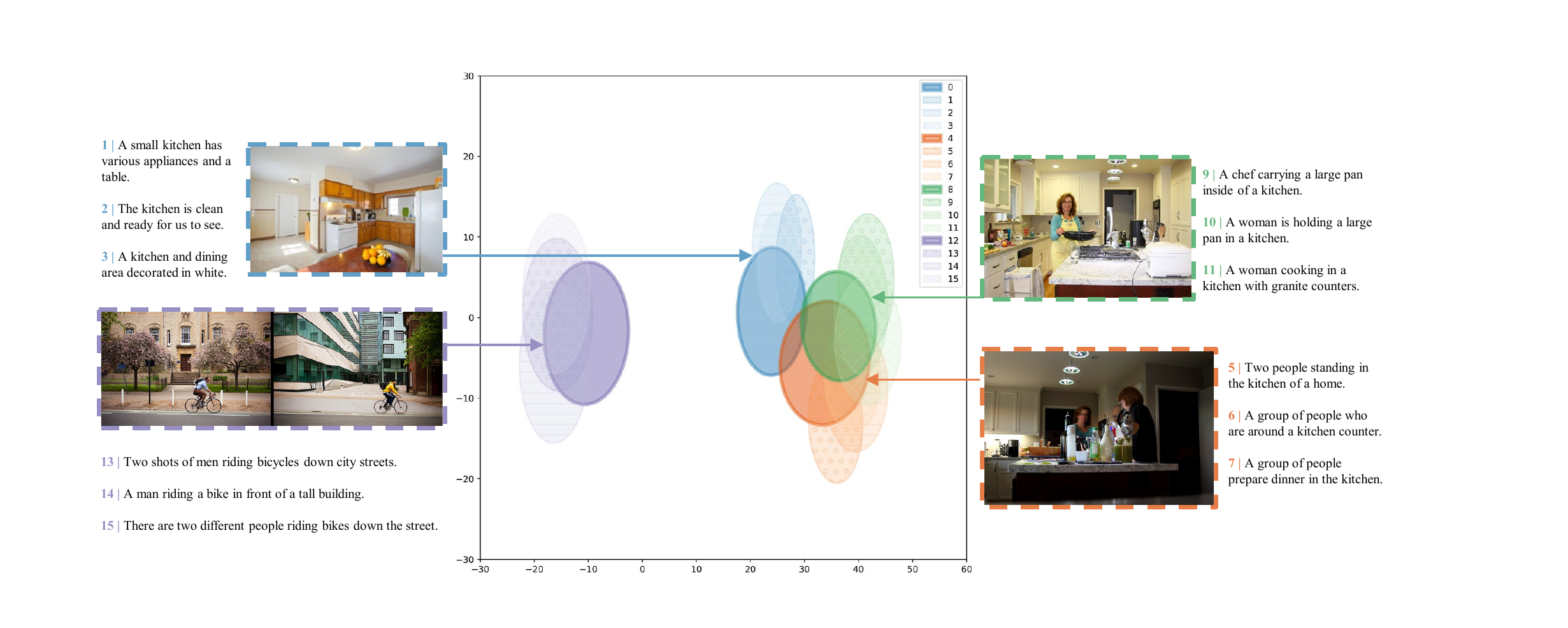}
  \caption{Additional example 3. There are some images and captions of ``woman'', ``kitchen'', ``street'', ``bike'' and so on.}
  \label{fig:visualization_others_3}
\end{figure*}

\subsection{Details and additional examples of visualization}
\label{append:add_examples_visualization}

After exacting the distribution representations from PDE, we conduct several 2D toy experiments by using clustering algorithms in machine learning. 
We utilize the pre-trained MAP with PDE to embed images and text onto distribution representations first. 
Then, the toy experiments are deployed to find non-linear connections from the input high-dimensional data. 
In detail, we consider the $\mu$ and $\sigma^2$ representations in the experiments separately and each experiment calculates more than a thousand image-text pairs. 
Figures~\ref{fig:visualization_others_1},~\ref{fig:visualization_others_2},~\ref{fig:visualization_others_3} shows several additional visualization examples of the distribution representations in different scenarios\footnotemark[4].

\footnotetext[3]{All images and related captions come from MSCOCO dataset~\cite{DBLP:conf/eccv/LinMBHPRDZ14/mscoco}.}

\subsection{ Visualization between point representations and distribution representations}

\begin{figure*}[tp]
\centering
  \includegraphics[width=\textwidth]{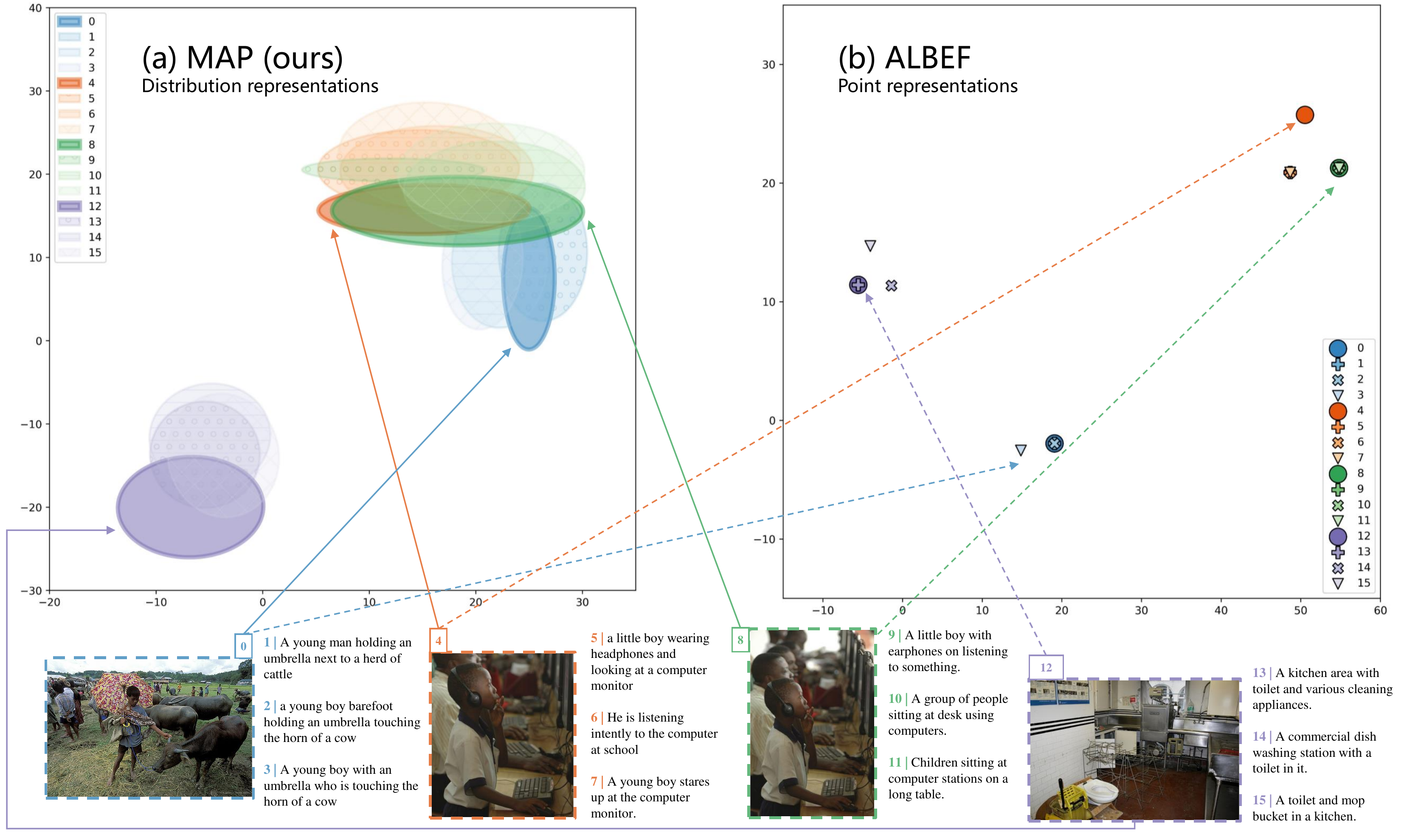}
  \caption{Visualization analysis on distribution representations and point representations.}
  \label{fig:visualization_analysis}
\end{figure*}

To explore the differences between representations, we compare our method with ALBEF.
For ALBEF (4M), we follow the same method and visualize the features of the same image-sentence pairs (see \cref{fig:visualization_analysis}).
Compared to ALBEF, our method takes advantages in capturing rich semantics and concepts in these pairs.

\section{Ethical Considerations}
\label{append:ethical}

Multimodal representation learning is a widely used technique that can have ethical effects.
Social bias seems to be rooted in the data due to accumulated biases on the web, such as gender bias in MSCOCO~\cite{DBLP:journals/corr/abs-1912-00578/mscoco_bias}.
We believe that our framework could be corrupted, leading to bias concerns, such as having preferences towards certain groups or features.
Given that the above problems cover a wide range of issues, such as privacy, fairness, and bias~\cite{alwahaby2022/evidence,hakami2020/learning_analytics}, we suggest applying our models to specific contextualization examples.
Users should also provide open discussions in their specific research areas and industrial environments.

\end{document}